\documentclass[iicol,sn-mathphys-ay]{sn-jnl}
\setcitestyle{aysep={,}} 
\let\cite\citep       

\usepackage{graphicx}%
\usepackage{multirow}%
\usepackage{amsmath,amssymb,amsfonts}%
\usepackage{amsthm}%
\usepackage{mathrsfs}%
\usepackage[title]{appendix}%
\usepackage{xcolor}%
\usepackage{textcomp}%
\usepackage{manyfoot}%
\usepackage{booktabs}%
\usepackage{algorithm}%
\usepackage{algorithmicx}%
\usepackage{algpseudocode}%
\usepackage{listings}%

\usepackage{subcaption} 

\theoremstyle{thmstyleone}%
\theoremstyle{thmstyletwo}%

\theoremstyle{thmstylethree}%

\begin{document}

\title[Model-Based Adaptive Precision Control for Tabletop Planar Pushing Under Uncertain Dynamics]{Model-Based Adaptive Precision Control for Tabletop Planar Pushing Under Uncertain Dynamics}


\author[1]{\fnm{Aydin} \sur{Ahmadi}}\email{aahmadi22@ku.edu.tr}
\author*[1]{\fnm{Baris} \sur{Akgun}}\email{baakgun@ku.edu.tr}

\affil[1]{\orgdiv{Computer Engineering Department}, \orgname{Koç University}, 
\orgaddress{\street{Rumelifeneri Yolu}, \city{Istanbul}, \postcode{34450}, 
\state{Istanbul}, \country{Turkey}}}


\abstract{Data-driven planar pushing methods have recently gained attention as they reduce manual engineering effort, and improve robustness and generalization compared to analytical approaches. However, most prior work targets narrow capabilities (e.g., discrete contact-point selection, precision, or single-task training), limiting broader applicability. We propose a model-based framework for non-prehensile tabletop pushing that leverages a single learned dynamics model to solve multiple tasks without retraining. Our approach uses a Gated Recurrent Unit (GRU)-based architecture with additional nonlinear layers to capture complex object–environment interactions while maintaining stable predictions. A tailored state–action representation enables generalization across uncertain dynamics, varying push lengths, and diverse manipulation goals. For control, we couple the learned dynamics with a modified sampling-based Model Predictive Path Integral (MPPI) controller, which generates adaptive, task-oriented actions. Multiple tasks are addressed simply by modifying the controller’s objective, without retraining. This framework supports side switching, variable-length pushes, and objectives such as precise positioning, trajectory following, and obstacle avoidance. Training is conducted in simulation with domain randomization to facilitate sim-to-real transfer. We first evaluate the architecture through ablation studies, demonstrating improved prediction accuracy and stable rollouts. We then validate the full system in simulated and real-world experiments using a Franka Panda robot with markerless tracking. Results show high success rates in precise positioning under strict thresholds and strong performance in both trajectory tracking and obstacle avoidance. We then extend our system to specialize for a single object type by 
training the model over a wider range of push lengths and designing an object-specific sampler that greatly reduces the number of steps to reach distant pose goals. We further show that the system has some object-generalization capabilities by testing it on two unseen object types.
}



\maketitle

\section{Introduction}\label{intro}

Pushing, a form of non-prehensile manipulation, lets robots pose objects without the need for grasping, allowing them to perform a variety of tasks. Various approaches have been explored to address this problem through planning, control, and reinforcement learning \citep{10.3389/frobt.2020.00008}.

Traditional approaches model push dynamics analytically and use a controller and/or a planner to generate actions \citep{mason1986mechanics, goyal1989limit, doi:10.1177/027836499901800504, lee2015hierarchical}. Modeling requires explicit  parameters of the objects, such as mass, center of mass (COM) and shape, along with environmental factors like friction and potential disturbances. However, modeling and controller design require sophisticated engineering. Additionally, they struggle with generalization in real-world scenarios when the objects or environment parameters change, often necessitating fine-tuning of parameters or even re-engineering. Moreover, these parameters are not always available, requiring system identification.

As a response to the limitations of traditional model-based methods, data-driven approaches have emerged to address some of the associated challenges \citep{bauza2017probabilistic, Li2018PushNetDP, bauza2019omnipush, kloss2020accurate, cong2020self, 9287951}. These methods are typically based on learning from experience, where a robot interacts with objects to collect data. The data is used to learn dynamics of object pushing which is then coupled with a push controller or to learn a pushing policy directly.
Rather than relying on explicitly specified physical parameters such as mass or friction, some of these approaches handle the inherent uncertainty and variability in real-world interactions through learning.

Methods based on model learning, model-based approaches, aim to approximate the forward dynamics of pushing through supervised learning. In this framework, the robot first collects a dataset of interaction trajectories, often consisting of proprioceptive or visual observations and corresponding object motion outcomes. A predictive model, typically a neural network or a probabilistic regressor, is then trained to map observations and actions to future object states. Once trained, this model is used in conjunction with a controller or a planner to compute action sequences that achieve a desired goal \citep{bauza2017probabilistic, Li2018PushNetDP, bauza2019omnipush, kloss2020accurate, cong2020self, 9287951}. One major advantage of this model-controller/planner decoupling is modularity: a well-learned dynamics model can be reused across multiple tasks, allowing us to plug in different controllers or planners depending on the specific objective. Moreover, these approaches are capable of learning end-to-end mappings from raw sensory inputs, such as vision or proprioception, to predicted outcomes, effectively bypassing the need for handcrafted features or explicit modeling of object properties. However, learned models are not without their limitations, such as computational cost of planning and potential out-of-distribution predictions.
 
Another subset of data-driven approaches is model-free methods. These approaches aim to directly learn a pushing policy that maps sensory inputs to actions, without explicitly modeling the underlying object dynamics. Given a goal state, the agent learns through trial and error to generate actions that guide the object toward the desired configuration. Once a policy is trained, action inference is extremely fast, requiring only a forward pass through a neural network, making model-free methods attractive for real-time applications. However, these methods require a lot of training hours for each desired task \citep{cong2022reinforcement, wang2023learning, ferrandis2023nonprehensile, dengler2024learning}.  

\begin{figure}[t]
\centering
\scriptsize
{\includegraphics[width=0.7\columnwidth]{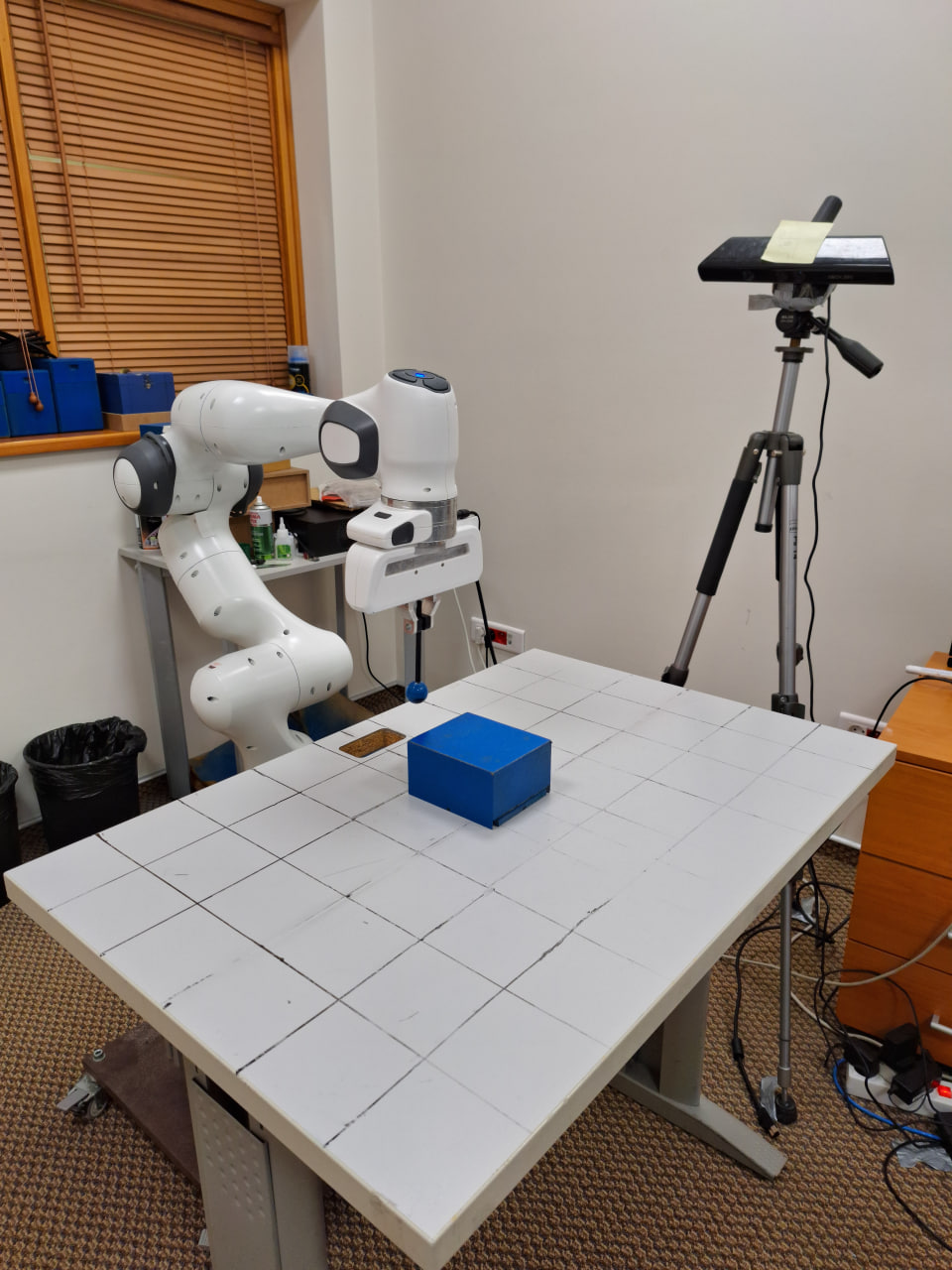}}

\caption{The real-robot setup. The robot holds a stick with a small ball at the end to interact with the objects. An RGBD camera is used for perception.}
\label{fig:setup}
\end{figure}

We adopt a model-based approach motivated by training economy and deployment benefits; train a push model once and adapt the controller/planner for multiple tasks. Towards this end, we propose a novel GRU-based architecture for learning push dynamics and use a sampling-based Model Predictive Control (MPC) approach, specifically a modified version of Model Predictive Path Integral (MPPI \citep{7487277}), for planar object manipulation on a table. We design our architecture to implicitly capture, potentially non-stationary object and environment dynamics by utilizing a recent set of previous state-action pairs, with an emphasis on the current state-action pair. However, our main aim is to develop a complete pushing system rather than developing the best possible model.

Our unique state-action representation includes the previous change of object pose, current object relative robot position (contact position) and the object relative push vector (both magnitude and direction). Our state-action representation and our choice of sampling-based MPC allow for changing the object sides during pushing. In addition, our sampling approach is goal-aware, which lets us generate different length pushes based on the current goal and adapt the samples for the task at hand. To the best of our knowledge, this is the first model-based approach that employs multi-length pushes. While the majority of existing methods consider only the direction of pushes with a fixed magnitude, our approach explicitly accounts for both direction and magnitude, enabling more flexible strategies. Our approach handles local optima for positioning, allows for additional objectives such as obstacle avoidance or different tasks such as trajectory following, and can trade off task duration with precision. We train our model in simulation and collect different length and direction pushes with domain-randomization \citep{tobin2017domain, peng2018sim} to facilitate sim-to-real transfer. We evaluate our approach both in simulation and on a real-world  setup as shown in Fig.~\ref{fig:setup}. In addition, we develop an additional object-specific sampling procedure to improve control and test it with a model trained on more data. This tailor-made controller is designed to show the modularity of our approach and its adaptability to specific situations.

The contributions of our work are:
\begin{itemize}
    \item A novel adaptive non-prehensile pushing system that integrates a specifically designed and novel state-action representation with a dedicated forward dynamics model for predicting pushing outcomes, requiring no explicit physics parameters and generalizing across tasks and a set of objects without task-specific tuning. Our main novelty lies particularly in the combination and interaction.

    \item Modified, goal-aware and object-specific MPC based controllers that can trade off task duration and task precision (in both translation and rotation axes), and allow variable-length pushes, side switching, and different task objectives.
    
    

    \item Detailed evaluations both in simulation and on a real robot setup using Franka Emika Panda robot, demonstrating the effectiveness of our method.
\end{itemize}

\section{Related work}\label{related work}
Pushing-based manipulation can be divided into (i) analytical methods, (ii) model-based data-driven methods, and (iii) model-free data-driven methods. In the first two, models for push dynamics and controllers/planners are used to generate actions. The model is engineered in the first and learned in the second. In the third, a push policy is directly learned.

\subsection{Analytical methods}
\citet{mason1986mechanics} develops an analytical model for quasi-static planar pushing.
\citet{goyal1989limit} expand this with the limit surface idea, which defines the relationship between sliding motion of the object and friction under the quasi-static assumption. This is used to determine the object velocity for a given pushing force. \citet{219921} investigates planar pushing using sliding and sticking motions, and employs a search algorithm for planning. \citet{lee2015hierarchical} introduce a hierarchical planning method for generating sequences of non-prehensile actions, where contact points for pushing are derived as part of the planning process. This approach is designed to reduce the overall search space by decomposing the full manipulation task into smaller sub-problems, allowing more efficient planning in complex environments. While analytical approaches offer interpretability and task-specific optimality under idealized conditions, their practical application is constrained by the need for accurate object-specific parameters such as mass, center of mass, and surface friction. These values are often difficult to measure in real-world scenarios, and even small deviations can lead to substantial performance degradation. They also lack generalization and are resource-demanding.

\subsection{Data-driven model-based methods}
Data-driven model-based methods aim to reduce engineering effort and better handle uncertainty. \citet{arruda2017uncertainty} use Gaussian Process Regression and an Ensemble of Mixture Density Networks to evaluate uncertainty of object motion predictions. While their method demonstrates the ability to switch object sides during planning, it lacks evaluation in sim-to-real scenarios, limiting its generalizability. 
Additionally, their approach does not support multi-length pushes and does not include planning-based task evaluations. Push-Net  leverages an LSTM module to encode temporal pushing interactions and incorporates an auxiliary objective to estimate the object's center of mass, implicitly embedding physical reasoning \citep{Li2018PushNetDP}. This approach uses object-segmented masks as visual input and demonstrates the ability to generalize to novel objects. 
However, their method relies on a rather loose success criterion and uses a single fixed push length, which limits its precision and contributes to a noticeably lower success rate reported in their evaluations.

\citet{yu2016more, bauza2017probabilistic, bauza2019omnipush} demonstrate the effectiveness of learning models based on collected data that include shape, friction, mass distribution, and perceptual information. \citet{yu2016more} present a large-scale dataset of planar pushing interactions, comprising over one million samples that include both positional data of the pusher and object, as well as corresponding interaction forces. \citet{bauza2017probabilistic} use a Gaussian process-based model to predict both the outcome and variability of planar pushing, incorporating pusher velocity to relax the quasi-static assumption. While effective, their approach offers limited generalization across variations in physical properties, as it neither incorporates domain randomization nor leverages strategies to enhance robustness beyond the training distribution. 

More recently, \citet{bauza2019omnipush} present a large dataset containing object pose information and visual data (RGB-D), and also introduces models trained on this dataset. 
However, their approach struggles with precise control of cubic objects, as it relies on randomly selected push points over a broad area and applies a single, fixed push length. These limitations are clearly reflected in their reported results. \citet{kloss2020accurate} introduce a hybrid model that integrates deep learning and analytical physics to predict planar object motion using RGB-D data. Their method estimates object pose and friction parameters through an analytical filter and learns an affordance model via CNNs. It uses object masks to classify motion outcomes, blending perception with physical modeling. While it requires no prior object knowledge, its property estimations are limited and error-prone, and it relies on assumptions of uniform properties and known friction, which limits generalization.
 
There are model-based methods that do not rely on object-environment parameters. \citet{cong2020self} propose a model-based strategy using a recurrent model and sampling-based MPC for control, which is similar to our approach. However, their approach ignores object orientation control, suffers from high sampling costs, does not consider multi-length pushes, and does not include additional tasks. Another drawback of their method is that the direction and pose of the first actions do not affect the action selection for the subsequent steps.
Additionally, it exhibited a low success rate even under relatively relaxed success criteria, achieving below 90\% success at large threshold values. \citet{9287951} introduce a few-shot learning approach for predicting the motion of objects with varying shapes. While their method demonstrates promising results for translation prediction, it overlooks critical factors such as object orientation and the historical trajectory of object motion, both of which are essential for achieving smooth and reliable rearrangement. In contrast, our model addresses these limitations by incorporating both historical actions and environmental dynamics, enabling accurate predictions of object motion in terms of both position and orientation. Furthermore, our predictions demonstrate higher accuracy compared to the statistical performance reported in their work. 

More recently, \citet{10802843} proposed an online framework for trajectory-following tasks by leveraging a non-parametric model, showing promising results for both model learning and control. However, their approach relies on single-step inverse dynamics learning to construct trajectories, which limits generalization and makes real world  deployment time-consuming and error-prone. In contrast, our system is built around  state-action pairs that capture the full interaction history, enabling a unified  framework capable of achieving multiple tasks, including goal-reaching, trajectory following, and obstacle avoidance, without retraining or task-specific tuning. Furthermore, while their perturbed inverse dynamics formulation does not support obstacle avoidance or broader task objectives, our sampling-based MPC naturally accommodates any task for which an objective function can be defined. Trained entirely in simulation with domain randomization, our system requires less than three hours from data collection to controller deployment.

\subsection{Model-free methods}
Model-free reinforcement learning (RL) based methods overcome the online scalability challenges of MPC, at the expense of extensive training. \citet{cong2022reinforcement} employ an RL method incorporating visuo-proprioception for objects of various shapes in planar pushing tasks. However, their performance is lacking and they do not incorporate object orientation in their objective. \citet{wang2023learning} propose a model-free RL method that uses contact force in their rewards. They only use a positioning task even though they utilize object and end-effector orientations. \citet{ferrandis2023nonprehensile} introduce a RL-based method for pushing cubic objects, using categorical exploration.

While their method shows promising results, there are several drawbacks. The required amount of transitions for proper training is in the order of billions. Since they use global coordinates for observations, their operational table size and target location are limited. A later version \citet{dengler2024learning} handles object collisions, however, as with any task-specific model-free method, the policies must be retrained for each task. Our system can be trained and deployed within hours, generalizes across multiple objectives without retraining, and can easily trade-off task duration and precision without additional training, which are not achievable with model-free methods.

\section{Push Dynamics Model}
We use a neural network to represent our push dynamics model. Its inputs are the recent history of object-relative robot poses and the relative robot motions, the current object-relative robot pose and the relative robot motion to be executed. Its architecture is designed to infer the physical parameters via the historical inputs while focusing on the latest action for accurate predictions. Its output is the predicted relative-object pose change. We train it in simulation with domain randomization to facilitate sim-to-real transfer. In this section, we explain the details of this model. 
In addition to our base model, we further train an additional model to be used with an object-specialized controller which we also discuss here.

\subsection{State-Action Representation and Model Architecture}\label{sec:repr_arch}

\begin{figure}[t]
\renewcommand{\arraystretch}{0.5}
\centering
\scriptsize
\includegraphics[width=0.8\columnwidth]{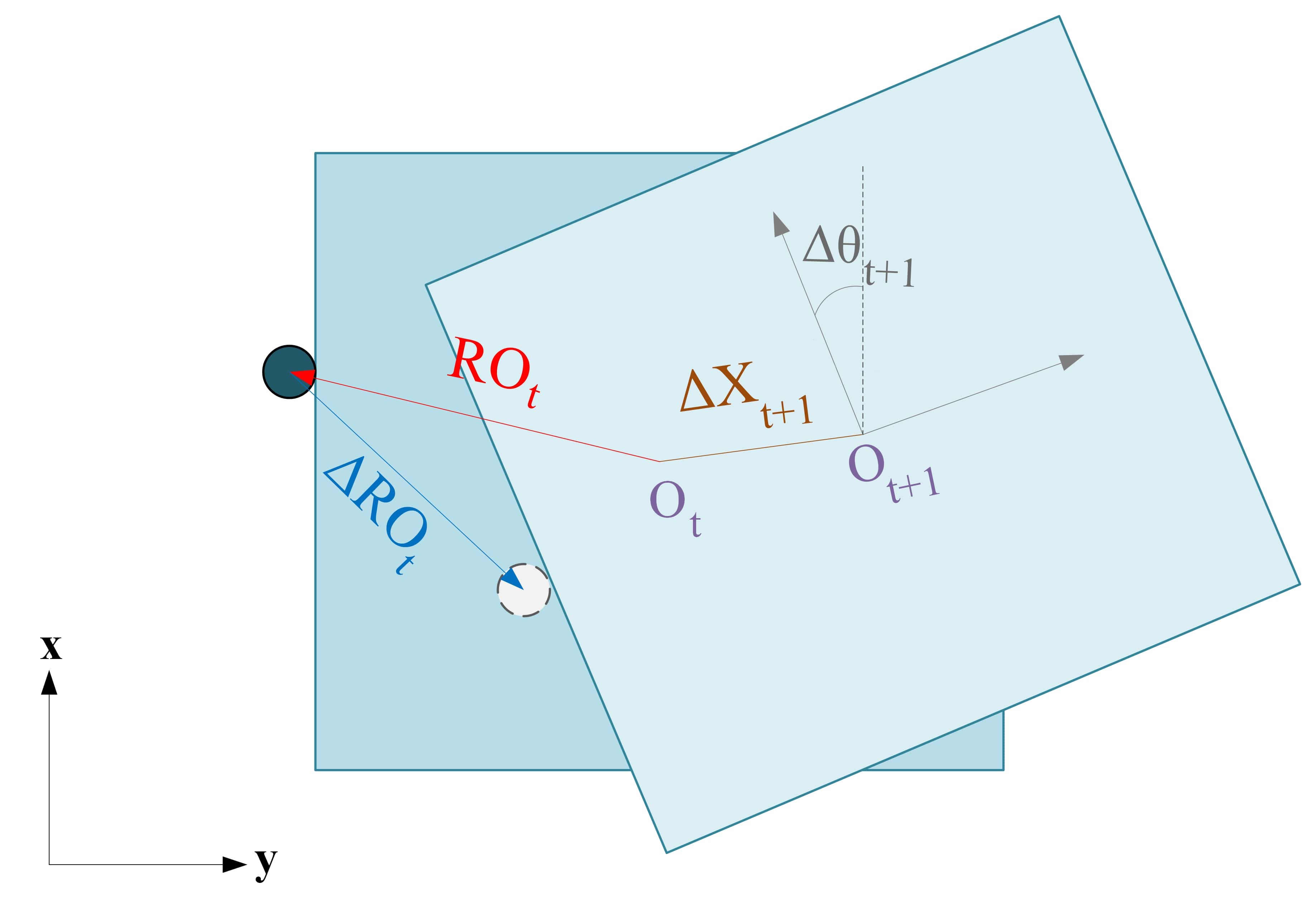}
\caption{State-action representation. $O_t$ represents the 2D object pose before the  push and $O_{t+1}$ represents it after the push. $RO_t$ represents the position of the pusher with respect to object before the push and \( \Delta RO_t \) represents the pusher motion. \( \Delta X_{t+1} \) and \( \Delta \theta_{t+1} \) denote the changes in the object's position and orientation, respectively, after the push, and $\Delta O_{t+1} = (\Delta X_{t+1}, \Delta \theta_{t+1})$.}
\label{fig:state_rep}
\end{figure}

Our state-action representation includes relative object-robot configurations and robot actions. We include a recent history of these to implicitly capture the environmental parameters (e.g. friction, mass etc.). The end-effector of the robot is assumed to be a point so we only include position. However, our predicted output is the change of object pose, including both position and orientation in the pushing plane. 

The single time-step versions of the representation components are geometrically depicted in Fig.~\ref{fig:state_rep}. Let $t$ be the current time step, $O_t$ the object pose in the plane and $R_t$ the 2D pusher position at $t$.  Then, the state-action tuple includes the following three; (1) relative object motion from the previous time step ($\Delta O_t = O_{t-1}^{-1} O_t$), (2) the relative position of the robot pusher with respect to the object ($RO_t = O_{t}^{-1} R_t$) and (3) the current relative pusher motion ($\Delta RO_t = R_{t+1}-R_{t}$). Then, the current state-action representation is the history from time step $t-w+1$ to $t$ (where $w$ denotes the window length), represented as:

\begin{equation}
H_t = \{(\Delta O_{k}, RO_{k}, \Delta RO_{k})\}_{k=t-w+1}^{t}
\end{equation}

\begin{figure*}[t]
\renewcommand{\arraystretch}{5}
\centering
\scriptsize
\includegraphics[width=\textwidth]{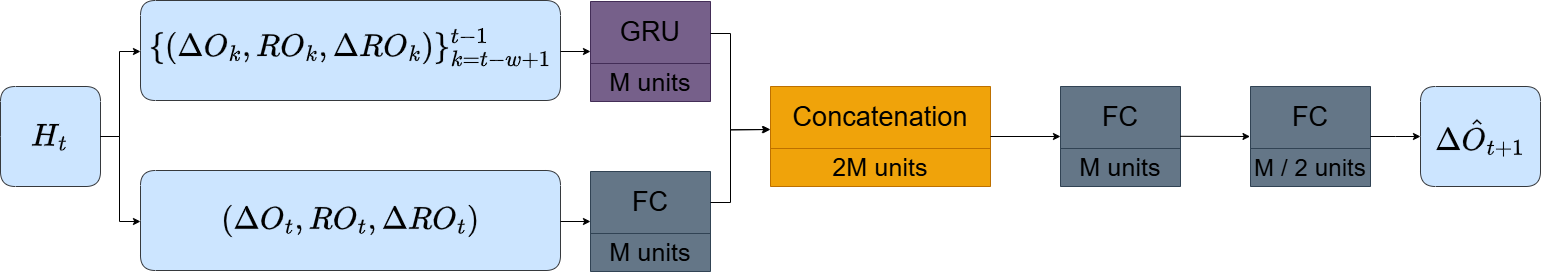}
\vspace{2pt}
\caption{Push dynamics model architecture. The ($RO_t$, $\Delta RO_t$) parts of the input constitute the action.}
\label{fig:model}
\end{figure*}

Our model, $f(\cdot)$, takes $H_t$, without any explicit object-environment physical parameters, and predicts the relative change of object pose, $\Delta \hat{O}_{t+1}$ ($\Delta \hat{O}_{t+1} = f(H_t)$). The architecture of our model is depicted in Fig.~\ref{fig:model}. The history up to the previous time-step (from $t-w+1$ to $t-1$) is input to Gated Recurrent Unit (GRU) layers to implicitly infer the environment's physical parameters and the most recent tuple is separately fed to fully connected layers to improve prediction accuracy. The FC and GRU output dimensions are the same to avoid one of them dominating the other. The outputs of these layers are then concatenated and passed through two more fully connected layers before the output layer. All FC layers have ReLU activations. From the perspective of the model, the action part is the location, direction and magnitude of the push ($RO_t, \Delta RO_t$) and the previous pushes are part of the state. 


Under the quasi-static assumption, we model each push at the displacement level. The direction and magnitude of the pusher motion are encoded by $\Delta R_O$, while pusher velocity and action duration are not included as independent inputs to the learned dynamics model. Consequently, the current model is intended for quasi-static pushing and is not expected to generalize to dynamic pushing regimes in which inertial or speed-dependent effects become significant.

\subsection{Data Collection} \label{sec:datacollection}
\begin{figure}[t]
\renewcommand{\arraystretch}{0.5}
\centering
\scriptsize
\includegraphics[width=0.8\columnwidth]{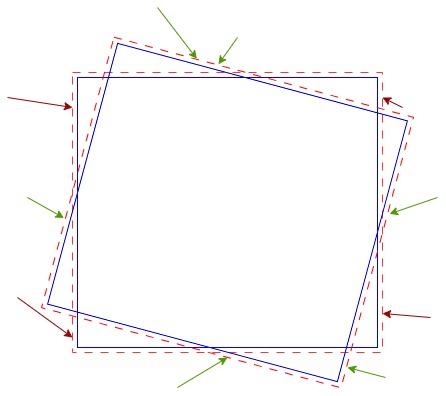}
\vspace{2pt}
\caption{Data collection process. This figure illustrates examples of individual pushes. Actions are sampled with directions uniformly distributed between $-45^\circ$ and $45^\circ$, and push magnitudes ranging from 2\,mm to 3\,cm. The red dots indicate the boundary of relative poses where actions are applied, designed to prevent premature contact (i.e., touching) before the action is executed. Arrow colors are associated with different objects
}
\label{fig:data_collection_draw}
\end{figure}

To train our model, we collect training data through random pushing interactions in a simulated environment using PyBullet, building upon the framework provided by \citet{ferrandis2023nonprehensile}. A total of 10,000 episodes are recorded, each consisting of 100 time-steps (i.e., pushes) applied to a planar cubic object with randomized size and physical parameters. To ensure diversity in object poses and avoid bias in the training distribution, we incrementally rotate the object by one degree across episodes. The pushing magnitudes range from 2\,mm to 3\,cm, and push directions are sampled uniformly within a symmetric cone of angle $\pi/4$ oriented towards the object at the point of contact. This setup ensures a broad distribution of contact angles and motion outcomes. The pushes are applied from points sampled on the perimeter of a square surrounding the object, with a margin of 2\,mm beyond its actual size, assuming the plane normal (z-axis) remains fixed. Fig. ~\ref{fig:data_collection_draw} illustrates a sample pushing configuration from the dataset. 

Our data collection, and the system as a whole, operates within the quasi-static regime, a ubiquitous assumption in non-prehensile planar pushing~\citep{mason1986mechanics, 10.3389/frobt.2020.00008}. In this regime the pusher moves slowly enough that inertial effects are negligible relative to friction, so the object moves only while being pushed and comes to rest as soon as pushing stops.

To enhance the generalization of our learned model and facilitate sim-to-real transfer, we employ domain randomization during data collection. Following the principles established in \citet{tobin2017domain, peng2018sim}, we randomly vary several environmental and object parameters at the beginning of each episode. These parameters include the object’s mass, dimensions, friction coefficient, and the restitution of the supporting surface. The exact ranges and details of these parameters are listed in Table~\ref{tab:domain_randomization_params}. This strategy enables the model to learn robust contact dynamics across a wide variety of scenarios, reducing overfitting to a narrow domain. Notably, we observed that the model trained entirely on simulated data generalized effectively to real-world robot experiments without requiring any additional fine-tuning. This highlights the effectiveness of combining fully random pushing actions with domain randomization for robust and scalable data-driven manipulation learning.

\textbf{Data for Object-Specific Controller:} The data collection procedure is identical apart from the fact that we sample pushing magnitudes between 3\,mm and 5\,cm. 

\begin{table}[t]
\centering
\caption{Domain randomization parameters}

\scriptsize
\begin{tabular}{c c}
\hline
Parameter & Sampling Distribution \\
\hline
Friction & \( [0.5, 0.7] \) \\
Restitution & \( [0.4, 0.6] \) \ \\
Object Length & \( [0.11, 0.13] \, \text{m} \) \\
Object Width & \( [0.09, 0.11] \, \text{m} \) \\
Object Mass & \( [0.3, 0.7] \, \text{kg} \) \\
\hline
\end{tabular}
\label{tab:domain_randomization_params}
\end{table}

\subsection{Training}
We use two separate models for predicting the 2D translational motion and the rotational motion about the plane-normal. We have empirically observed that this configuration performs better in practice. We posit that this is mainly due to removing the need for tuning the relative weights of the position and the orientation losses. 

The loss for the translation prediction model is defined as the mean squared distance between the target and the predicted translations. The loss for the orientation prediction model is defined as the mean non-oriented angular difference between the target and the predicted orientations. The non-oriented angle is calculated as follows:
\begin{equation}
D_{\theta}(\theta_1, \theta_2) = \min (| \theta_{1} - \theta_{2} |, \;\; 2\pi - | \theta_{1} - \theta_{2} |)
\label{eqn:orn_loss}
\end{equation}

The models have the exact same architecture other than the output size. We use $M=64$ (see Fig.~\ref{fig:model}) to instantiate both of them. We perform a 0.7, 0.2, 0.1 train-validation-test set splits. The models were trained using the Adam optimizer \citep{kingma2014adam} with a learning rate scheduler. A total of 40 epochs were enough for convergence in all cases and we take the model with the best validation performance.

\section{Control}
\label{subsec:control}
\subsection{Base Controller}
We use a modified version of the MPPI controller which is a sampling-based MPC \citep{7487277}. The basic idea is to sample a number of actions, predict corresponding rollouts using our learned model, calculate their objective value (e.g. cost, reward), calculate the next action based on these, apply the first action, then repeat until success or the horizon limit is reached. Note that the objective functions are specific to tasks and we assume they are to be minimized. For example, the distance of the last position of a rollout to the target can be an objective function for positioning. 

The main MPPI control loop is given in Alg.~\ref{algorithm:mppi_outer} which is fairly straightforward. The main effort is in the line 6 with the \textsc{ActionSelection} function. The base version for our state-action representation and model is given in Alg.~\ref{algorithm:mppi_base_as}. For each rollout of a given horizon length, the algorithm samples actions (line 9), calculates the change in object pose with the learned dynamics model autoregressively (line 10), updates the current pose, history, and current rollout pose and action trajectories (lines 11-14). Each rollout's objective value is calculated when the horizon is reached (line 16). Then after all rollouts are calculated, the final action is selected based on calculated objectives and the sampled actions (line 20). For compactness, $f(y,RO,\Delta RO)$ at line 10 denotes constructing the current model input from the preceding state-action history $y$ and the candidate action $(RO,\Delta RO)$, consistent with the representation defined in Section~\ref{sec:repr_arch}.

We use a goal-aware sampling procedure in our base MPPI controller (line 8 in Alg.~\ref{algorithm:mppi_base_as}). We first sample a relative pusher position ($RO$) from our dataset used to learn the model. This is followed by a random push direction within a symmetric $\pi/4$ cone towards the object. We then randomly pick a push length but the upper limit is selected based on the distance to the goal. The direction and length is then used to calculate $\Delta RO$. This approach allows us to switch object sides and perform different length pushes. Note that for the base version of our controller, we always sample the same action during the rollouts (i.e., use the first sampled action $T$ many times back to back).





\begin{algorithm}[!t]
\caption{MPPI Control Loop}
\begin{algorithmic}[1]
\small
\State \textbf{Input:} Initial object pose $O_0$, learned dynamics model $f(\cdot)$, objective function $o(\cdot)$
\State \textbf{Parameters:} Number of samples $N$, rollout horizon $T$

\While{task objective not satisfied}
    \State Observe current object pose $O_t$
    \State Update state-action history $\bar{H}_t$
    \State $a_t \gets$ \textsc{ActionSelection}$(O_t, \bar{H}_t, N, T, f, o)$ 
    \State Execute action $a_t$ on the robot
    \EndWhile
\end{algorithmic}
\label{algorithm:mppi_outer}
\end{algorithm}

\begin{algorithm}[!t]
\caption{MPPI Action Selection}
\begin{algorithmic}[1]
\small
\State \textbf{Input:} Current object pose $O_t$, State-Action History $\bar{H}_t$, learned dynamics model $f(\cdot)$, objective function $o(\cdot)$, Number of samples $N$, rollout horizon $T$
\State \textbf{Output:} The next action ($RO_{t},\Delta RO_t$) 

\For{Rollouts $k = 1 \ldots N$}
    \State $x \gets O_t$ 
    \State $y \gets \bar{H}_t$
    \State $\tau_{k} \gets \{x\}$
    \State $a_k \gets \emptyset$
    \For{Time step $h = 0 \ldots T-1$}
        \State $RO,\Delta RO \gets SampleAction(x)^*$
        \State $\Delta \hat{O}_{h} = f(y, RO,\Delta RO)$
        \State $x \gets x + \Delta \hat{O}_{h}$ 
        \State $y \gets updateHist(y,\Delta \hat{O}_{h},RO_{k,h},\Delta RO_{k,h})$ 
        \State $\tau_{k} \gets \tau_{k} \cup x$
        \State $a_k \gets a_k \cup (RO,\Delta RO)$
    \EndFor
    
    \State $c_k \gets o(\tau_k)$
    \EndFor
    \State $C \gets c_{k=1\dots N}$
    \State $A \gets a_{k=1\dots N}$
    \State $RO_{t},\Delta RO_{t} \gets calcAction(C,A)$ 
\end{algorithmic}
\footnotesize{$^*$ For the base controller, this sampling step is performed only at the first rollout step and the sampled action is repeated for the remaining steps. For the object-specific controller, an action is sampled at each rollout step as described in Section~\ref{sec:obj_cont}.}
\label{algorithm:mppi_base_as}
\end{algorithm}

The action selection (line 20 in Alg.~\ref{algorithm:mppi_base_as}) is done by taking the weighted combination of the sampled actions. Individual action weights are calculated with the Eq.~\ref{eqn:norm_sample_cost}. Then, the actions are combined using the Eq.~\ref{eqn:action_select} where $a_t = [RO_t,\Delta RO_t]$ is the selected action. 

\begin{equation}
\omega_i = \frac{\exp\left(\frac{-c_i}{\lambda}\right)}{\sum_{j=1}^N \exp\left(\frac{-c_j}{\lambda}\right)}
\label{eqn:norm_sample_cost}
\end{equation}

\begin{equation}
a_t = \frac{\sum_{i=1}^N \left(\omega_i \times a_i\right)}{\sum _{i=1}^N \omega_i}
\label{eqn:action_select}
\end{equation}

This approach prioritizes actions that are more effective in reducing the objective value (e.g. distance) while still allowing other actions to affect the outcome to avoid being overly greedy. The trade-off is set by the parameter $\lambda$. However, the resulting relative pusher position may be in collision with the object. In that case, we restart the sampling process. If we cannot find a feasible solution in 20 restarts, we select the action with the lowest cost. 

To execute the action, the robot end-effector pose is calculated based on the current object pose $O_t$ and the selected relative pusher position $RO_t$. We use a motion-planner to reach this pose. Then, the robot is moved by $\Delta RO_t$, using a Cartesian controller. After the robot motion, the object pose is recalculated using an external camera. This process repeats until an objective (e.g., object within a threshold) or a maximum number of iterations are reached.

Recall that the push magnitude is selected based on distance to goal. For the base method, we adopt a two-stage approach. In the first stage, the primary aim is to rapidly bring the object close to the target which is achieved by sampling longer pushes.
We switch to the second stage once the object is within 1 cm of the target. In the second stage, the strategy changes to sampling shorter pushes. The idea is to perform precise adjustments and accurately pose the object. In the first stage, the pushes are sampled in the range 8mm-3cm, and in the second stage, from 1 mm to 8 mm.

\subsection{Object-Specific Controller} \label{sec:obj_cont}
We introduce an object-specific controller in addition to our base controller. The idea is to show the flexibility of our method and to explore performance limits when the target objects are constrained, which is a potential use case in controlled environments such as factories. The object-type we use for this controller is a rectangular prism. In our controller, we use the model trained with the data described at the end of Sect.~\ref{sec:datacollection}. 

Towards this end, we introduce several modifications aimed at enhancing both the diversity of sampled actions and the precision of the action selection process. First, we increase the number of sampled actions per control iteration, enabling a denser exploration of the action space and significantly increasing the likelihood of identifying high-quality actions. Instead of sampling $RO_t$ and $\Delta RO_t$ jointly, we now sample each component separately. A balanced number of $RO_t$ values are sampled from all four sides of the object, which encourages more diverse and symmetric interaction strategies. This is where the object-type comes into play; We use the object shape knowledge to sample. Note that this sampler is parametric and can accommodate different sized rectangular prisms. We also expand the sampling range of push magnitudes to span 3\,mm to 5\,cm, accommodating both coarse repositioning and fine-tuning within a unified framework.

Another key enhancement concerns the temporal structure of actions during the rollout. Rather than applying the same action across all steps in a rollout, we introduce directional variation at each step. Specifically, while the push magnitude remains constant, the direction changes within a range of $\pm5^\circ$ relative to the previous direction, other than the first one. We use two different initial ranges to get two different controller instances; narrow interval ($[-5^\circ, 5^\circ]$) for more direct pushes, or from a wider interval ($[-45^\circ, 45^\circ]$) for more deviated pushes toward the object.
Subsequent actions are then perturbed basedc on this initial value. This change allows the controller to explore more realistic and continuous action trajectories.

Finally, instead of selecting the action based on a weighted average over sampled rollouts, we adopt a 
greedy strategy by directly choosing the action with the minimum rollout cost. This avoids the dilution effect of averaging, which can lead to $RO_t$ values occasionally falling inside the object boundary. These modifications retain the overall structure of the controller while significantly improving sample efficiency, control accuracy, and robustness across a wide range of pushing scenarios. Ablations examining the effect of side-switching capability, the object-specific controller modifications relative to the base controller, and key hyperparameter choices are presented in Section~\ref{sec:ablations}.

%
%
%
%
%
%
%
%
%
%

\subsection{Tasks}
In our evaluations, we use precise posing of an object as our main task. The aim is to push the object to a target pose (position and orientation). We additionally consider two tasks; (i) trajectory following and (ii) obstacle avoidance. Trajectory following involves pushing the object along a series of waypoints.
Obstacle avoidance requires the system to reach a target pose while circumventing obstacles that may be on the direct path between the initial and target poses. Each task has its own objective function. 

We define the pose difference in Eq.~\ref{eqn:posediff}, where $P_i = [X_i, \theta_i]$ is the object pose, $X_i$ is the associated 2D position vector, $\theta_i$ is the associated orientation around the 2D plane normal, $D_\theta$ is from Eq.~\ref{eqn:orn_loss}, and $w_\theta$ is the orientation weight. 
\begin{equation}
D_{pose}(P_1,P_2) = \left\| X_{1} - X_{2} \right\|^2 + w_\theta D_\theta(\theta_{1},\theta_{2})^2
\label{eqn:posediff}
\end{equation}

\textbf{Precise Posing}: In this task, we define the objective function of a rollout as the pose difference between the predicted object pose of the last rollout time step and the target pose, as defined in Eq.~\ref{eqn:cost1}, where $\hat{P}_{\text{last}}$ is the predicted object pose at the last rollout time step.
\begin{equation}
c_{posing}(\tau_i) = D_{pose}(P_{\text{target}}, \hat{P}_{\text{last}}^i)
\label{eqn:cost1}
\end{equation}

We  have an alternative formulation which penalizes the robot travel between pushes, to prevent frequent side-switching for small gains. This is given in Eq.~\ref{eqn:cost2}, where $\phi$ is the motion planning penalty weight, $R_{\text{current}}$ is the last pusher position and $R_{\text{sampled}}$ is the sampled pusher position. 
\begin{equation}
c_{penalty}(\tau_i) = c_{posing}(\tau_i) + \phi \left\| R_{\text{current}} - R_{\text{sampled}} \right\|^2
\label{eqn:cost2}
\end{equation}

\textbf{Trajectory Following}: Trajectory following tasks may employ various objective functions, depending on the trajectory representation. In our case, we adopt a dense trajectory representation, meaning that waypoints are placed closely along the desired path of the object. Since we repeat the same push during rollouts, it is enough for us to look only at the first step. We also do not expect frequent side-switching due to dense waypoints.
As such, we use the simple objective defined in Eq.~\ref{eqn:cost3}, where $P_{\text{p}}$ is the next waypoint and $\hat{P}_{\text{first}}$ is the predicted object pose at the first rollout time step. 
\begin{equation}
c_{tf}(\tau_i) = D_{pose}(P_{\text{p}}, \hat{P}_{\text{first}}^i)
\label{eqn:cost3}
\end{equation}

\textbf{Obstacle Avoidance}: This task is similar to precise positioning with the added requirement of avoiding obstacles. There are multiple objective functions for obstacle avoidance in the literature. We chose a function that exponentially penalizes the distance between the object and the obstacles and the distance between the pusher and the obstacles as defined in Eq.~\ref{eqn:cost4}, where $d_o$/$d_p$  represents the closest distance between the predicted object surface/pusher position during the rollout steps and the closest obstacle surface.
\begin{equation}
c_{oa} = w_1 \exp\left( -\alpha_o \, d_o \right) + w_2 \exp\left( -\alpha_p \, d_{p} \right)
\label{eqn:cost4}
\end{equation}
The weights, $w_1$ and $w_2$, determine the relative importance of penalizing proximity between the object and the obstacles ($w_1$) and between the pusher and the obstacles ($w_2$). The parameters $\alpha_o$ and $\alpha_p$ control the sensitivity of the exponential penalty to the distances $d_o$ and $d_p$, respectively. In our experiments we use $w_1 = 10$, $w_2 = 10$, and $\alpha_o = \alpha_p = 100\ln(10)$. During this task, the obstacle avoidance objective is added to the positioning objective only if $d_o$ or $d_p$ is less than 1 cm. This formulation penalizes actions that bring the object or the pusher closer than 1 cm to any obstacle, encouraging the selection of safer paths while still making progress towards the target.

\section{Experiments and Results}
\label{sec:experiments_results}
\subsection{Model Evaluation}

We evaluated our model on one-step predictions from the test set.
We pick two baselines to demonstrate the effectiveness of utilizing the history while focusing on the current time-step in our architecture. The first baseline is the 2-layer LSTM model proposed by \citet{cong2020self}, which we use directly with the same number of hidden units as in our model. The second baseline is a model that does not incorporate history.

Table~\ref{tab:model_comparison} presents the prediction performance of these three models and the model for the object-specific version. We further divided the test set into two sets to capture the performance of the models with different length pushes. The "big" set contains pushes from 8 mm to 3 cm, while the "small" set contains pushes from 2 mm to 8 mm.  For the object-specific version the pushes are from 3mm to 5cm.

Our model outperforms both of the baselines. The performance compared to the no history baseline is expected. The no history model cannot infer the physical properties with just looking at the current state. On the other hand, the first baseline performed the worst - particularly bad for position prediction - even though it was taking the history into account. 
One potential reason is that the action cannot influence the prediction enough and it gets dwarfed by the history. Since our dataset is heavily randomized, this particular architecture (limited inductive bias compared to ours) arguably did not have enough data points to generalize. As such, we conclude that our architecture provides a good balance between historical information to infer pushing properties and the current action to perform accurate predictions. Although the new model exhibits higher prediction errors compared to the Big Set baseline, it operates under more challenging conditions. Unlike prior models that are trained on restricted push magnitudes or separated datasets, our approach learns a single unified dynamics model from the full spectrum of pushing behaviors, including significantly larger pushes. Crucially, the model remains stable and consistent despite this expanded action space, demonstrating strong robustness and generality.


\begin{table*}[tb]
    \centering
    \scriptsize
    \small 
    \caption{Object motion prediction performance of our proposed model and baselines. 
    The \textbf{Avg. Pos. Err.} column denotes the position prediction mean error in millimeters with the standard deviation in parentheses.
    The \textbf{Avg. Rel. Err.} column denotes the position prediction average relative error (average error over average push length).
    The \textbf{Avg. Ort. Err.} column denotes the orientation prediction mean error in radians with the standard deviation in parentheses. The last row shows the results for the object-specific version model. Small Set includes pushes between 2mm and 8mm, Big Set includes 8mm to 3cm and the Larger Set includes 3mm to 5cm.}
    \vspace{3pt}
    \begin{tabular}{lccc} 
        \hline
        \textbf{Model / Set} & \textbf{Avg. Pos. Err.} & \textbf{Avg. Rel. Err.} & \textbf{Avg. Ort. Err.}  \\
        \hline
        No History - Small Set & 0.6 (0.4) & 0.12 & 0.004 (0.004)  \\
        \citep{cong2020self} - Small Set & 1.2 (0.6) & 0.24 & 0.016 (0.012) \\
        Ours - Small Set & \textbf{0.3 (0.3)} & \textbf{0.06} & \textbf{0.003 (0.003)} \\
        \hline
        No History - Big Set & 1 (1) & 0.045 & 0.01 (0.015) \\
        \citep{cong2020self} - Big Set & 6.2 (3.6) & 0.28 & 0.077 (0.052) \\
        Ours - Big Set & \textbf{0.8 (0.8)} & \textbf{0.036} & \textbf{0.007 (0.009)} \\
        \hline
        New model - Single Set  & \textbf{1.5 (1.7)} & \textbf{0.032} & \textbf{0.016 (0.023)}\\
        \hline
    \end{tabular}
    
\label{tab:model_comparison}
\end{table*}

\subsection{Simulation}
\begin{figure}[t]
\renewcommand{\arraystretch}{0.5}
\centering
\scriptsize
\includegraphics[width=0.5\columnwidth]{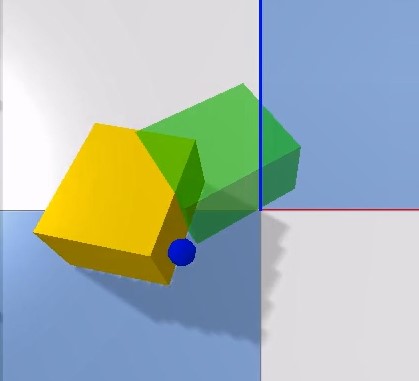}

\caption{Simulation environment. The figure shows a frame from the precision positioning task. The yellow cube represents the current object pose, the green marker indicates the target pose, and the blue sphere represents the pusher.}
\label{fig:sim_env}
\end{figure}

We use the same PyBullet simulation environment to collect model data to evaluate our approach, as illustrated in Fig.~\ref{fig:sim_env}. The environment consists of a cubic object to be pushed toward a specified goal pose using a spherical pusher of radius 1.25\,cm. The horizon length, $T$, for precise posing and obstacle avoidance is $5$, and it is 1 for trajectory following. The number of samples, $N$, is 20 per control step for precise posing and 100 for the other tasks. The orientation weight, $w_\theta$, is 0.025 for precise posing, 0 for trajectory following, and 0.03 for obstacle avoidance when the object is within 5 cm of the target pose and 0 elsewhere. The motion planning weight $\phi$ is dynamic and is the quarter of the distance to the target position in the precise posing task until the object is within 2 cm of the target, at which point it becomes zero. For obstacle avoidance, the robot weight remains zero but for trajectory following we have tested both zero and 0.01 weights. In the object-specific version of our controller, we introduce several key modifications. For precision positioning, we also use a rollout horizon of $T=3$ and sample $N=100$ actions, distributing them equally across all four sides of the object (25 samples per side). 

For benchmarking purposes, we adopt the model-free policy introduced by \citet{ferrandis2023nonprehensile} as a baseline for the precise posing task. Specifically, we use the policy trained and released by the original authors without further fine-tuning. This policy was developed using reinforcement learning in simulation and is designed to directly control the pusher to move the object to a desired pose. The observation space for this model-free policy includes both the current pose of the object and the pose of the pusher, providing the agent with sufficient spatial context to infer appropriate actions. The action space consists of velocity commands for the pusher in the 2D plane, allowing for continuous control over its motion.

In addition, we evaluate both of our proposed precise posing objectives based on Eq.~\ref{eqn:cost1} and Eq.~\ref{eqn:cost2} within our framework. These objectives are designed to explicitly guide the pushing behavior by incorporating goal pose alignment and, optionally, penalizing excessive pusher movement.

Beyond precise posing, we also apply our approach to other manipulation tasks, including obstacle avoidance and trajectory following. This broader evaluation demonstrates the generalizability and robustness of our model-based method across diverse pushing scenarios, where each task imposes different constraints on the controller and relies on distinct cost formulations.

\begin{figure*}[t]
\centering
\scriptsize
\includegraphics[width=\textwidth]{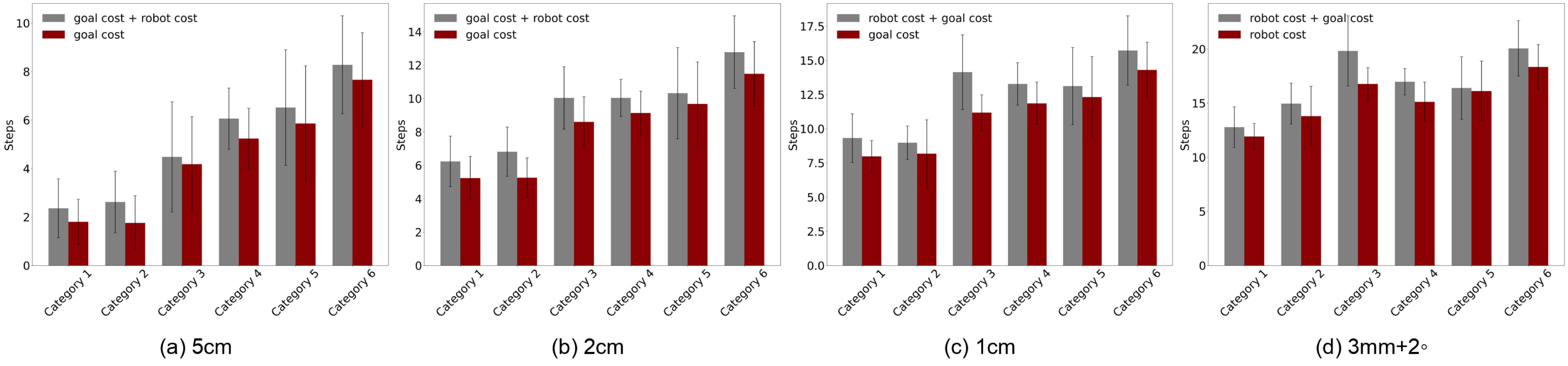}
\caption{Number of steps taken by the basic method across different categories in simulation. The subfigures correspond to different success thresholds: 5 cm, 2 cm, 1 cm, and 3 mm + 2 degrees from left to right. The vertical axes show the steps to reach these thresholds in simulation. The categories in the horizontal axes correspond to the Cartesian product of $\{(5cm-15cm),(15cm-20cm)\}$ (5cm-15cm),(15cm-20cm) distances and $\{(0^{\circ}–30^{\circ}),(30^{\circ}–60^{\circ}),(60^{\circ}–90^{\circ})\}$ orientation differences between the initial pose and the target pose (e.g. Category 1 is (5cm-15cm) and (0°–30°), 6 is (15cm-20cm) and (60°-90°)). The crimson and gray bars correspond to the objective functions defined in Eq.~\ref{eqn:cost1} and Eq.~\ref{eqn:cost2} respectively.}
\label{fig:threshold_steps_simulation}
\end{figure*}

\begin{figure}[t]
\renewcommand{\arraystretch}{0.5}
\centering
\scriptsize
\includegraphics[width=0.9\columnwidth]{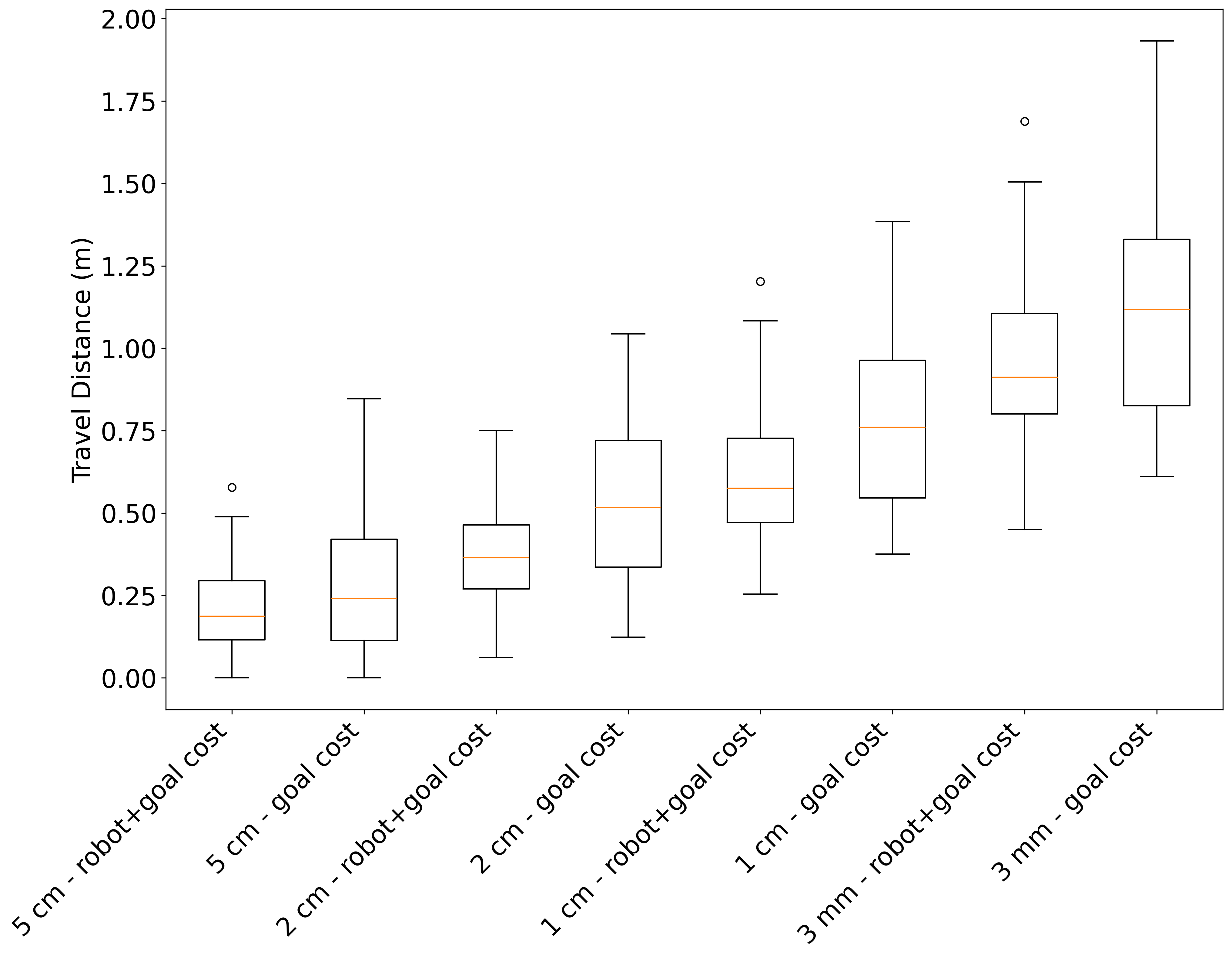}
\caption{The robot travel for our basic method in simulation. The "goal cost" corresponds to  Eq.~\ref{eqn:cost1} and "robot cost + goal cost" corresponds to Eq.~\ref{eqn:cost2}. The figure illustrates that pushing the object toward a distant goal with larger orientation changes results in increased robot travel.}
\label{fig:threshold_travel_simulation}
\end{figure}

\noindent \textbf{Precise Manipulation}: 
There are six categories of experiments, each comprising 10 cases. The experiments are organized into two primary categories based on the distance between the initial pose and the target pose. The first category involves target distances ranging from 5 cm to 15 cm, and the second category from 15 cm to 20 cm. Within each distance category, the experiments are further subdivided based on the target and the initial orientation differences into three intervals: 0°–30°, 30°–60°, and 60°–90°. Each individual experiment was repeated 5 times, for a total of $2 \times 3 \times 10 \times 5 = 300$ episodes.

The main results are shown in ~Fig. ~\ref{fig:threshold_steps_simulation}. The trends within each objective are expected. The number of required steps increase with threshold strictness, with target distance and with orientation difference. However, the model performance did not vary much between the (0°–30°) and (30°–60°) orientation difference categories.  Overall, fewer than 20 steps are required for the object to reach within 1 cm of the target pose. These results also demonstrate the advantage of decoupling the learned model from the controller: greater precision can be achieved simply by allowing more time, without necessitating retraining or altering the policy structure. Although a direct comparison with existing methods is challenging due to variations in evaluation setups and implementation details, our method achieves a favorable trade-off between efficiency and accuracy. It consistently requires few steps while maintaining high success rates under stringent error thresholds, attributable to the robustness of the learned dynamics and our approach to generating variable-length push actions.

Including the robot motion penalty increased the required steps as expected, since not switching sides may prolong posing. Fig. ~\ref{fig:threshold_travel_simulation} demonstrates that incorporating the motion planning cost reduces the robot travel which suggests that trading off number of steps with robot travel is feasible. The average time to finish the pushing episodes is 5.96 seconds.

\begin{figure*}[!t]
\centering
\scriptsize
\captionsetup{justification=centerlast}
\includegraphics[width=\textwidth]{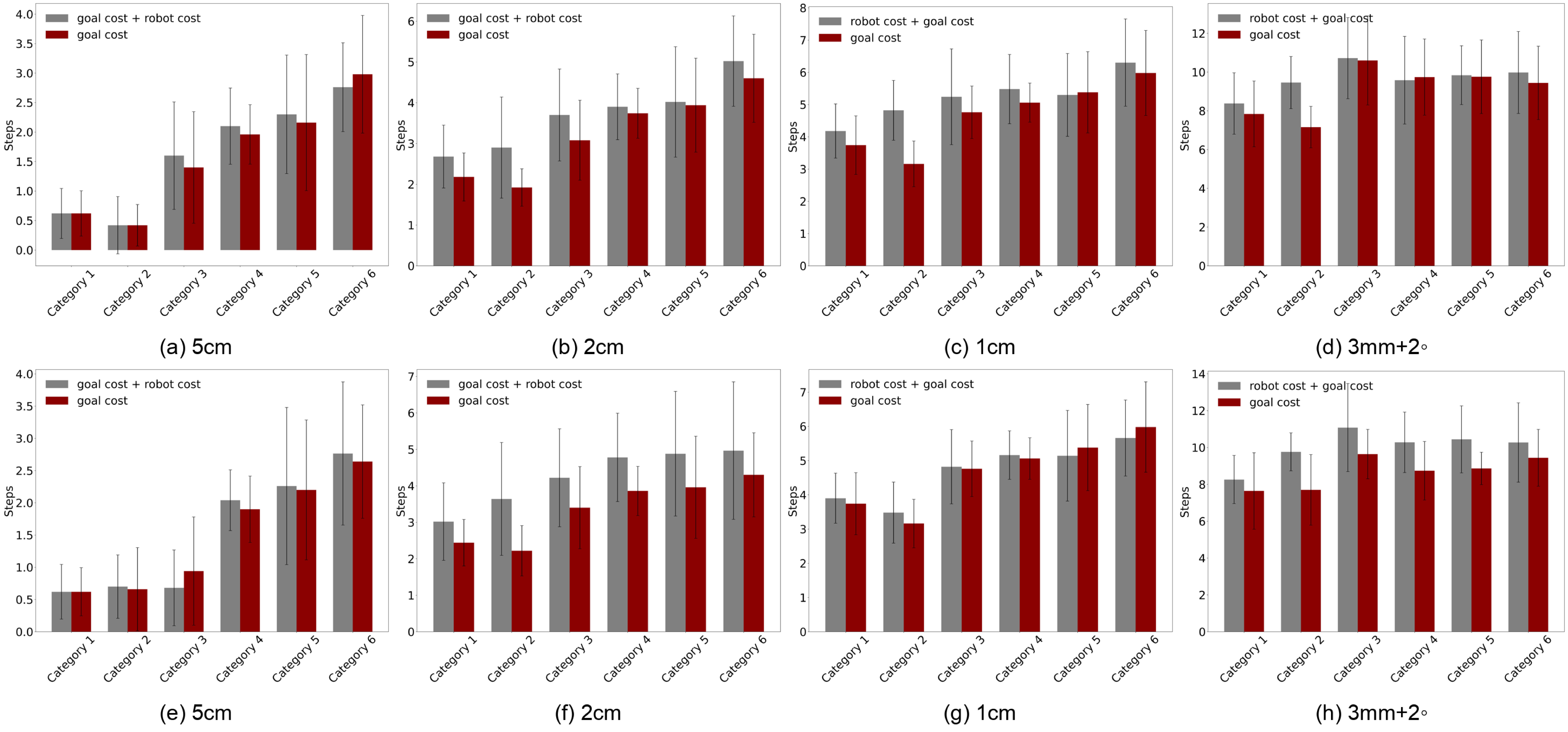}


\caption{Number of steps taken by the object-specific method across different categories in short-distance pushing scenarios in simulation with a horizon of 3. Results are evaluated at four success thresholds—5\,cm, 2\,cm, 1\,cm, and 3\,mm~$+~2^\circ$—shown from left to right in each group. Each subfigure reports the number of simulation steps required to reach the specified threshold. Rows correspond to different combinations of rollout horizon and initial push direction range: \textbf{Row 1} uses wide orientation sampling $[-45^\circ, 45^\circ]$; \textbf{Row 2} uses with narrow orientation sampling $[-5^\circ, 5^\circ]$. The horizontal axis within each plot categorizes the tasks based on the Cartesian product of translation distance intervals $\{(5\text{\,cm}–15\text{\,cm}),\ (15\text{\,cm}–20\text{\,cm})\}$ and orientation differences $\{(0^\circ–30^\circ),\ (30^\circ–60^\circ),\ (60^\circ–90^\circ)\}$. For example, Category 1 denotes 5–15\,cm distance and $0^\circ$–$30^\circ$ rotation, while Category 6 denotes 15–20\,cm and $60^\circ$–$90^\circ$.}
\label{fig:simulation_results_short_improved}
\end{figure*}
\begin{figure*}[!t]
\centering
\scriptsize
\captionsetup{justification=centerlast}
\includegraphics[width=\textwidth]{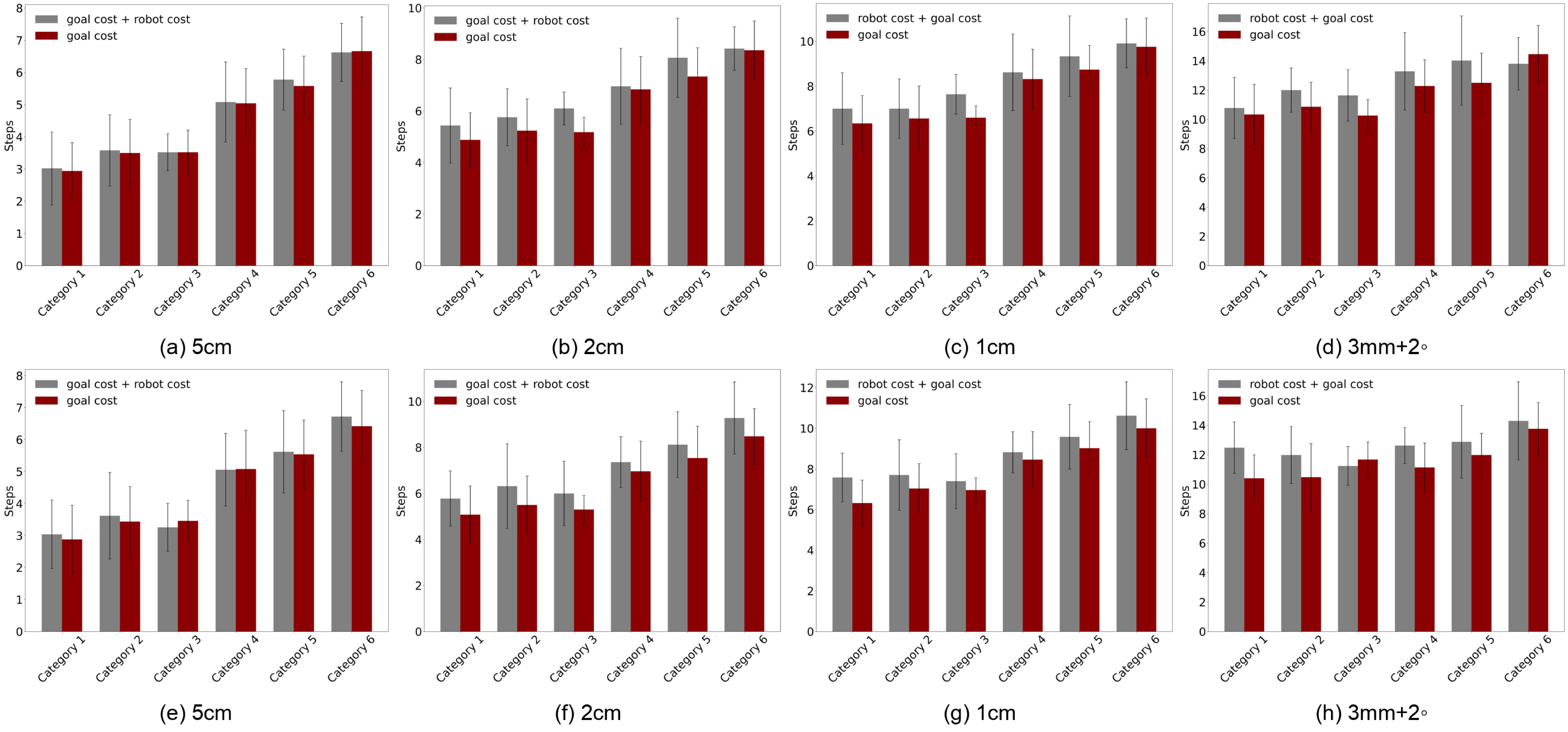}



\caption{Number of steps taken by the object-specific method across different categories in long-distance pushing scenarios in simulation with a horizon of 3. Results are evaluated at four success thresholds—5\,cm, 2\,cm, 1\,cm, and 3\,mm $+~2^\circ$—shown from left to right in each group. Each subfigure reports the number of simulation steps required to reach the specified threshold. Rows correspond to different combinations of rollout horizon and initial push direction range: \textbf{Row 1} uses wide sampling; and \textbf{Row 2} uses narrow sampling. The horizontal axis in each plot categorizes tasks based on translation distance $\{10\text{\,cm}–20\text{\,cm},\ 20\text{\,cm}–30\text{\,cm}\}$ and rotation difference $\{0^\circ–30^\circ,\ 30^\circ–60^\circ,\ 60^\circ–90^\circ\}$ between initial and target poses.}

\label{fig:simulation_results_distant_improved}
\end{figure*}

\begin{figure*}[!t]
\centering
\scriptsize
\captionsetup{justification=centerlast}
\includegraphics[width=\textwidth]{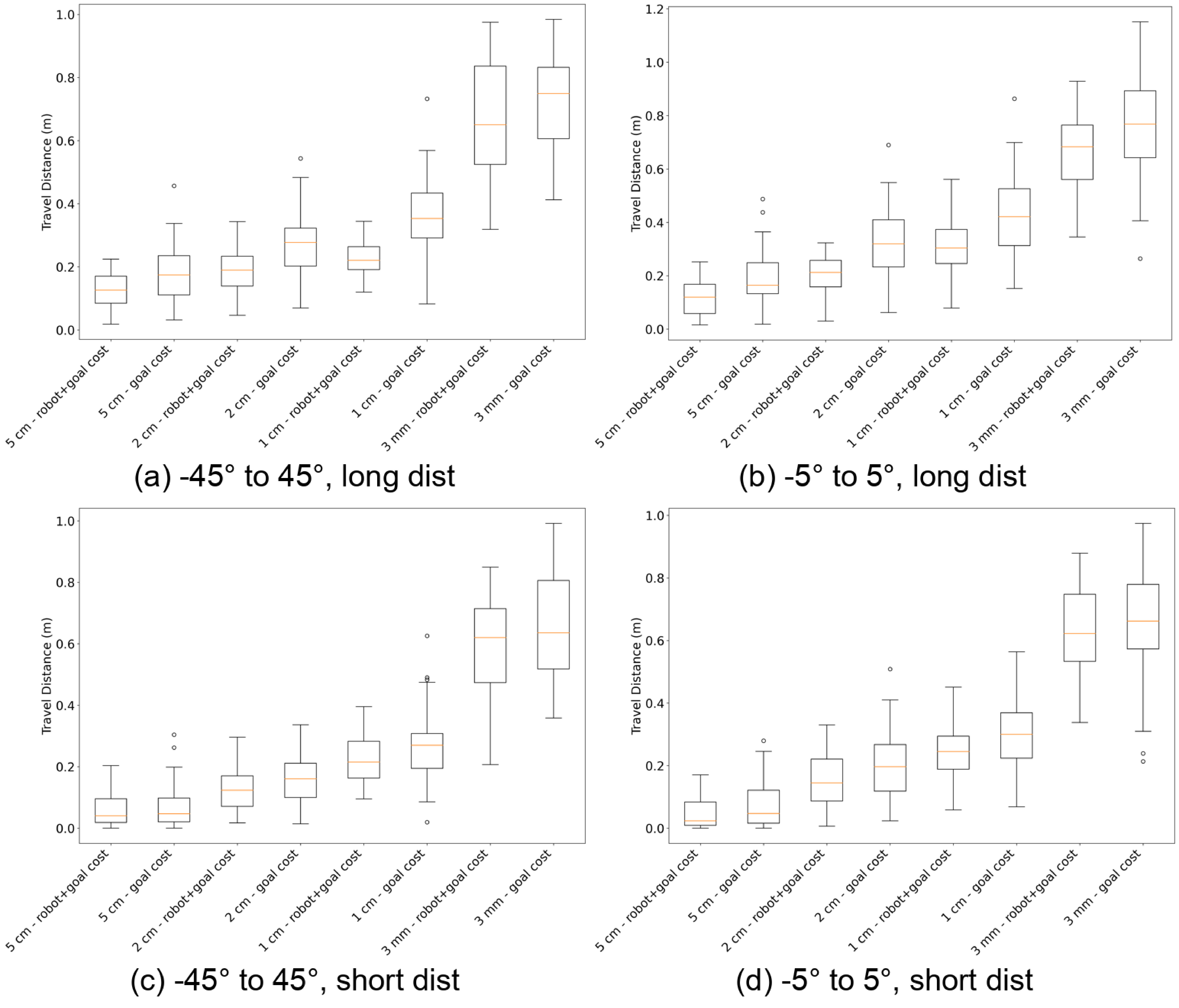}

\caption{Robot travel distance for the object-specific version in simulation. Comparison of ``goal cost'' (Eq.~\ref{eqn:cost1}) and ``robot cost + goal cost'' (Eq.~\ref{eqn:cost2}) across different pushing scenarios. Initial push directions are sampled from either a wide range ($[-45^\circ, 45^\circ]$) in \textbf{(a)}, \textbf{(c)} or a narrow range ($[-5^\circ, 5^\circ]$) in \textbf{(b)}, \textbf{(d)}. Each pair contrasts long vs.\ short pushing distances.}
\label{fig:combined_travel_plots}
\end{figure*}

We evaluated the object-specific controller on the same six categories described above, as well as an additional set of experiments involving larger target distances. Specifically, the new set consists of two distance ranges: $10--20\,\text{cm}$ and $20--30\,\text{cm}$. Each distance range is further subdivided into three orientation intervals: $0^\circ$--$30^\circ$, $30^\circ$--$60^\circ$, and $60^\circ$--$90^\circ$, resulting in six new categories. For all categories, we tested two sampling strategies for the initial push direction: a narrow range of $[-5^\circ, 5^\circ]$ and a wide range of $[-45^\circ, 45^\circ]$.  The results, presented in Fig.~\ref{fig:simulation_results_short_improved} and Fig.~\ref{fig:simulation_results_distant_improved}, show that the number of steps required to reach each success threshold is reduced by approximately half. Furthermore, Fig.~\ref{fig:combined_travel_plots} illustrates the substantial reduction in robot travel distance achieved by the object-specific controller. We conducted experiments with horizons of 3. A smaller horizon may increase the number of steps, but in our case horizon 3 is more suitable since the target is not far for pushes of up to $5\,cm$ 
per step considering the rollout length. On average, the episode completion time was $5.80\,seconds$ 
for horizon~3.

We have also compared our method with a model-free baseline. We show the aggregate results in Table~\ref{tab:rl_outs_comparison}. Our method achieves 100\% task completion rate for both of the thresholds. However, the model-free method performs poorly in addition to the sharp decline between thresholds. It can be argued that this baseline would also perform similarly given more training time. However, at some point, training becomes infeasible which is due to the nature of model-free methods. This demonstrates that for certain problems, model-based methods provide more flexibility. In fact, with extended operation (removing stopping thresholds) and careful hyperparameter tuning, we can achieve errors as low as 1~mm and 1° but the required number of steps make systematic evaluation difficult. 

\begin{table}[!t]
\centering
\scriptsize
\caption{Comparison of Our Method and Model-Free Baseline \citep{ferrandis2023nonprehensile}.}
\begin{tabular}{l c c} 
\hline
\textbf{Threshold} & \textbf{Ours} & \textbf{Model-Free} \\
\hline
$3~\mathrm{mm},~2^\circ$ & $1.000$ & $0.303$ \\
$7.5~\mathrm{mm},~5^\circ$ & $1.000$ & $0.886$ \\
\hline
\end{tabular}
\label{tab:rl_outs_comparison}
\end{table}

Lastly, we log the object pose predictions during our experiments. The average error between the model prediction and the object pose after taking the selected action was \( 1 \pm 0.84 \) mm for distance and \( 0.0075 \pm 0.0082 \) radians for orientation, similar to the "big set" model evaluation results for our basic method.

\begin{figure*}[!t]
\centering
\scriptsize
\includegraphics[width=\textwidth]{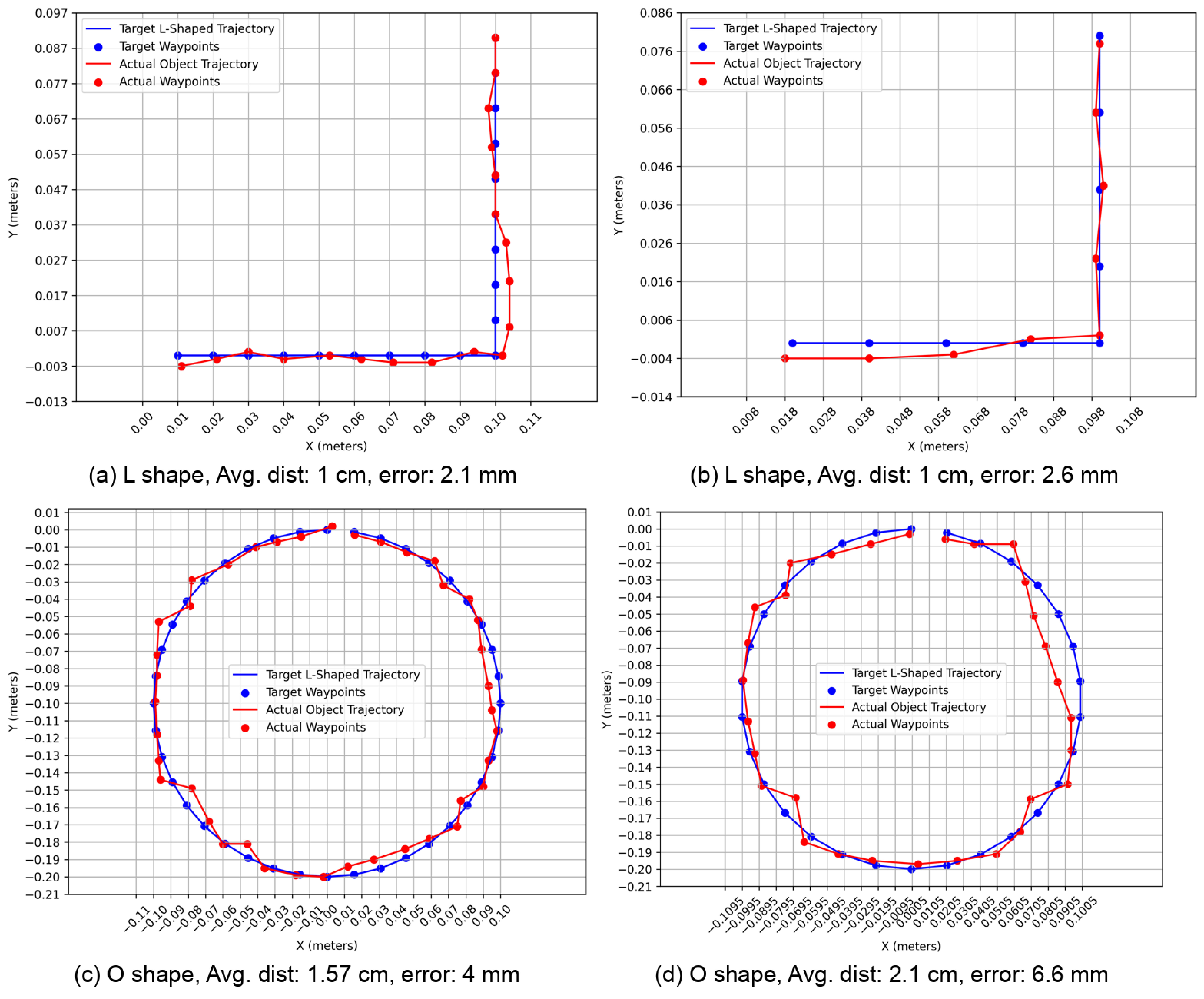}



\caption{Representative simulation results of trajectory following without orientation cost. 
Top row: L-shaped trajectories with push sizes of 1 cm and corresponding average position errors of 2.1 mm and 2.6 mm. 
Bottom row: circular (O-shaped) trajectories with push sizes of 1.57 cm and 2.1 cm, and corresponding average position errors of 4 mm and 6.6 mm. 
These cases demonstrate the controller's high accuracy across different path geometries and step sizes.
}
\label{fig:tfs}
\end{figure*}

\noindent \textbf{Trajectory Following}:
We conduct experiments using both circular and "L"-shaped trajectories composed of dense waypoints. Our controller uses Eq.~\ref{eqn:cost3} and performs single-step rollouts to decide the next action. Waypoints are switched once the object reaches sufficiently close proximity to the current target.
We tested circular trajectories with radii of 10 cm, 20 cm, and 30 cm, and "L"-shaped paths with side lengths of 10 cm, 20 cm, and 30 cm. For each shape and size, we evaluated two cases: one with no orientation objective and one with a small orientation cost of $0.01$. The average position error between the desired and the actual trajectories for following the circular shape ranged from 3 to 4 mm, while for the "L"-shaped trajectory, the average error was between 2 and 3 mm, with the average push length between 1.5 cm and 2 cm. These results highlight how the system's performance varies with different spatial separations between consecutive waypoints. Fig.~\ref{fig:tfs} illustrates representative examples of trajectory following performance in simulation.

\noindent \textbf{Obstacle Avoidance}: 
As mentioned before, this task is precise posing with an additional obstacle avoidance term, as defined in Eq.~\ref{eqn:cost4}. We use the objective function defined in Eq.~\ref{eqn:cost1} for posing so that the motion planning cost does not implicitly help obstacle avoidance with less side switching. 

\begin{figure*}[!t]
\centering
\scriptsize
\captionsetup{justification=centerlast} 
\includegraphics[width=\textwidth]{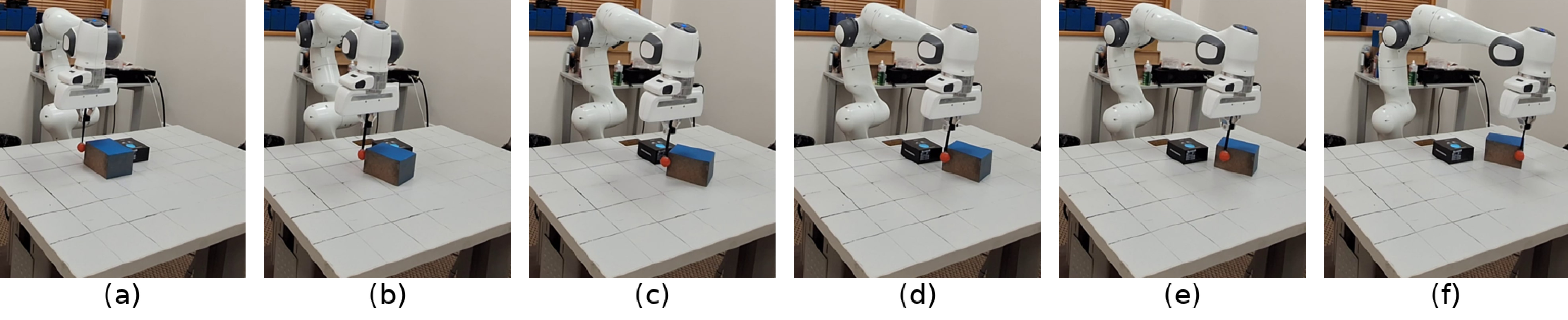}

\caption{Frames from real-world object avoidance case.}

\label{fig:real-world-case}
\end{figure*}

We use rectangular-prism objects of height 7cm and square base of 3cm and 5cm to simulate obstacles in the environment and evaluate our method under various obstacle configurations from single to multiple obstacles. We ensure each configuration has a collision-free path and repeat them 5 times. An example obstacle avoidance trajectory for the real robot is given in Fig.~\ref{fig:real-world-case}. Certain configurations require large orientation changes to traverse narrow passages. 

Configurations with single objects or two objects with relatively wide separation are typically easy to handle for the method. As the separation between obstacles decreases, the success rate drops. Additionally, the success rate decreases with the number of obstacles, as expected. Our approach struggles in scenarios requiring the object to rotate more than $60^\circ$ to pass through narrow gaps between two closely placed obstacles. These larger rotations highlight the need for more advanced planning capabilities, as the current system's limited planning horizon and fixed action sequences restrict its ability to adapt to complex, intricate maneuvers.

Due to the local nature of the MPC framework, the proposed obstacle avoidance method is not complete. For example, certain cul-de-sac configurations lead to inevitable failures because the controller lacks global re-planning capability.



\begin{figure}[tb]
\centering
\scriptsize
\includegraphics[width=0.7\columnwidth]{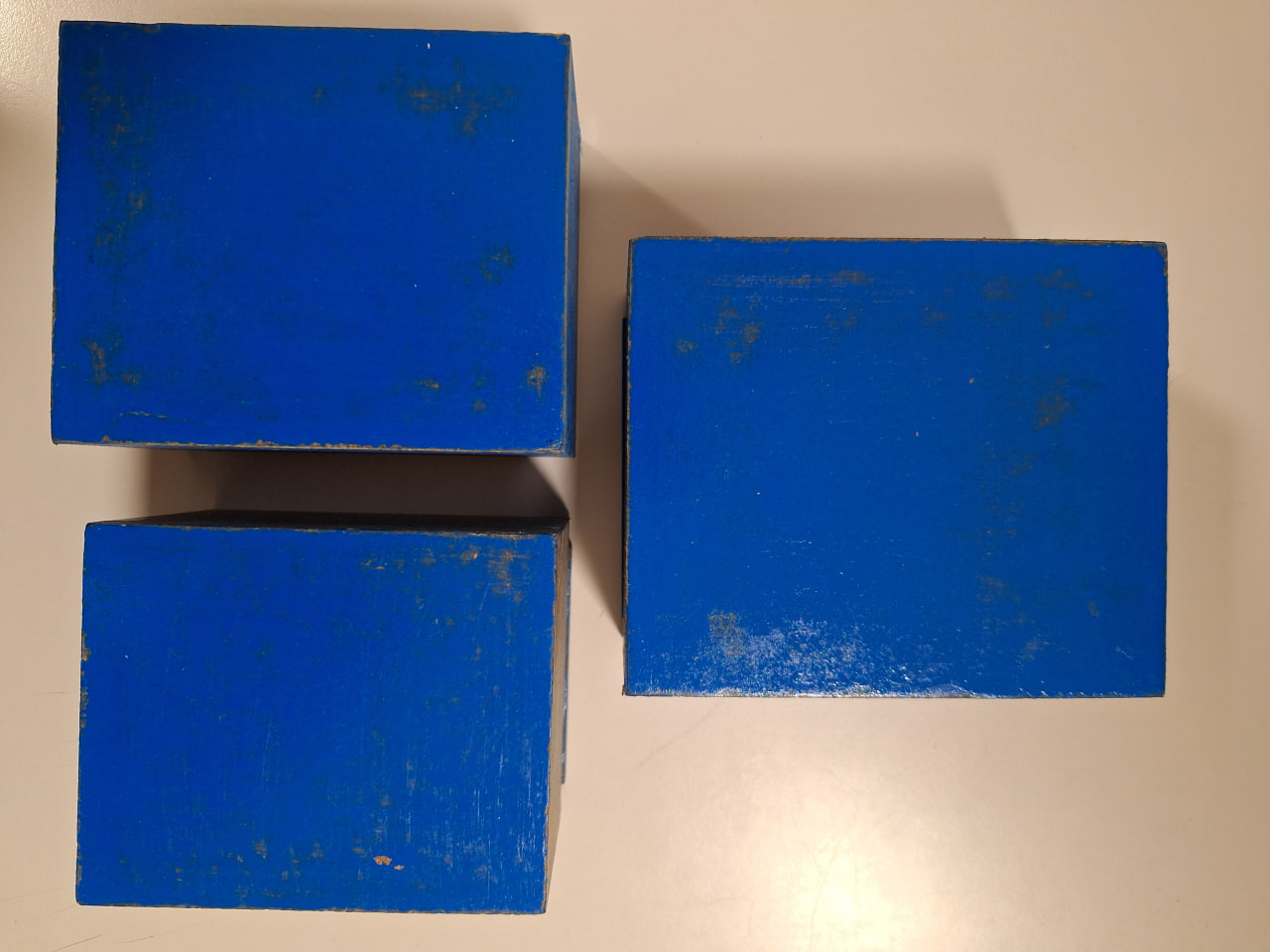}
\caption{The objects used in the real-world experiments. The box at the bottom, object 1, has dimensions of 9 \(\times\) 11 cm, the box at the top, object 2, has dimensions of 10 \(\times\) 12 cm, and the box on the right, object 3, has dimensions of 13 \(\times\) 11 cm.}
\label{fig:objects}
\end{figure}

\begin{figure}[tbp]
\centering
\captionsetup{justification=centerlast} 
\includegraphics[width=\columnwidth]{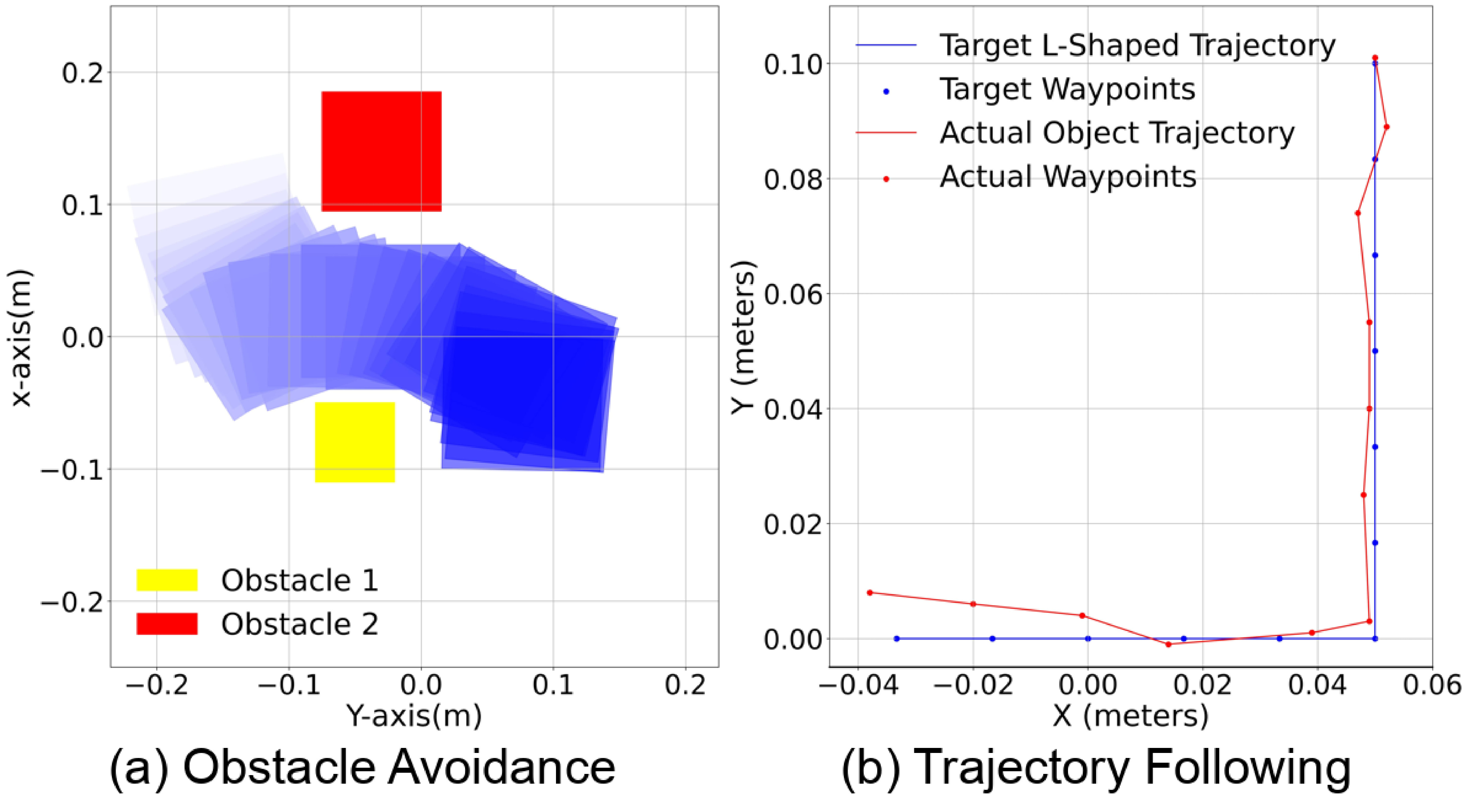}

\vspace{5pt}
\caption{\fontsize{8}{9}\selectfont Obstacle avoidance (a) and trajectory following (b) cases in the real world. (a) illustrates a two-obstacle scenario where the object rotates to pass through the narrow passage and then rotates again to reach the target pose. 
(b) depicts an "L"-shaped trajectory following in a real-world scenario, with a one-step average prediction error of 5.6 mm. 
}
\label{fig:real_world_tasks}
\end{figure}

\subsection{Real World}
\noindent \textbf{Real World Setup}:
We conduct a series of real-world experiments to evaluate the effectiveness of the basic planar pushing method. The robotic setup consists of a Franka Emika Panda robot, which holds a 15 cm stick equipped with an elastic spherical ball attached to its tip. A Kinect 360 RGBD camera is used to perform markerless tracking of object poses in real time. We use an old sensor without markers to have a challenging noisy environment to provide a sharp contrast to the simulation environment. The experiments were performed on a 90 cm by 60 cm table. The tracking standard deviation was around 1 mm and 1 degree but some far away parts of the table resulted in systematic errors due to camera angles which we avoid. This setup provides a realistic and challenging evaluation environment and can be seen in Fig.~\ref{fig:setup}. To ensure quasi-static conditions during real-world trials, the robot’s velocity is limited to a maximum of 10 cm/s. Unlike in simulation, where both velocity and step duration are explicitly defined, the real-world execution uses a simple position controller to push the object according to the selected action. To maintain accurate pose tracking, the robot takes a brief step backward after each action, allowing the perception system to reacquire the object pose before selecting and executing the next action. The speed of our real-world experiments is limited by the backward movement required at each step; however, we have demonstrated that our system operates efficiently and fast enough in simulation.

\noindent \textbf{Precise Manipulation}: 
We perform the precise posing task on three objects, shown in Fig.~\ref{fig:objects}. We divide the experiments into two categories: short distances (position: 9-15 cm, orientation: 0-60$^\circ$) and long distances (position: 15-20 cm, orientation: 60-90$^\circ$), with all objects starting at the same initial pose. We perform a total of 300 push episodes. Additionally, we conduct twenty experiments in which both the initial and target poses are randomly chosen. 

\begin{figure}[tb]
\centering
\scriptsize
\captionsetup{justification=centerlast} 
\includegraphics[width=\columnwidth]{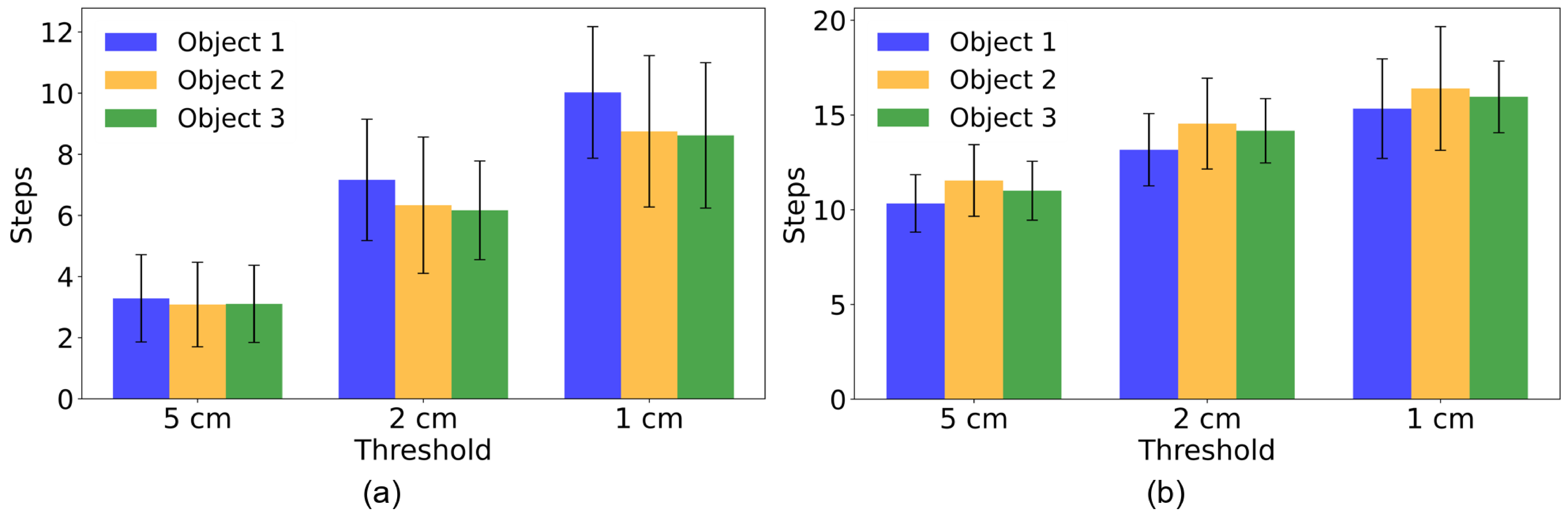}

\caption{The performance of our method in real-world experiments. The left subfigure corresponds to short distance (9-15 cm, 0°–60°) configurations and the right subfigure to long distance (15-20 cm, 60°–90°) configurations.
}

\label{fig:real_manipulation}
\end{figure}

Fig.~\ref{fig:real_manipulation} illustrates the number of steps required to reach the target poses. The results mirror that of the simulation case. In addition, we can see that the required number of steps increase with object size. Table~\ref{tab:real_control_per} shows the success rate of all the pushes for 1cm position and two a posteriori orientation thresholds. Note that although the objective function included orientation, the stopping condition did not, which explains the less than 100\% success rates. The numbers are fairly close for all cases and they are all relatively high. This shows the real-world performance of our approach. We lastly run 10 short distance object 1 cases with a (5 mm, 2$^\circ$) threshold with average 10.8 steps and 100\% success rate.

We also measured the one-step prediction error of the push dynamics model. The average distance error is \( 3.8 \) mm for the same starting pose cases and \( 4.8 \) mm for the random starting pose cases, while the average orientation error is \( 2.5^\circ \) for the aforementioned cases.

\begin{table}
\centering
\scriptsize
\caption{Success rates for reaching the target pose in manipulation tasks for less than the mentioned thresholds.}
\label{tab:real_control_per}

\begin{tabular}{c c c}
\hline
\textbf{Episode Type} & \textbf{5 degrees} & \textbf{10 degrees} \\
\hline
Long Distant Episodes & 0.875 & 0.98 \\
Short Distant Episodes & 0.86 & 0.97 \\
Random Episodes & 0.9 & 0.95 \\
\hline
\end{tabular}
\end{table}

\noindent \textbf{Trajectory Following}: 
The trajectory following evaluations are conducted with two distinct shapes for the object's path: a circular trajectory and an L-shaped trajectory. In each experiment, the robotic pusher was tasked with guiding the object along these predefined shapes while minimizing the distance between successive waypoints to less than half of the set distances. The average displacement error ranged from 5 to 7 mm, with the average push length between 1.67 cm and 2 cm for L-shaped trajectories. For circular trajectories, the average displacement error ranged from 6 to 7 mm, with the average push length again between 1.67 cm and 2 cm. Fig.~\ref{fig:real_world_tasks}b shows an example L-shaped trajectory following result.

\noindent \textbf{Obstacle Avoidance}: 
We evaluated the obstacle avoidance task through five experiments with a single obstacle and five experiments with two obstacles. In these tests, our method successfully reached the target pose while avoiding obstacles, maintaining a safe distance of at least 1 cm without any collisions. Fig.~\ref{fig:real-world-case} illustrates the robot pushing the object to the target pose while avoiding a single obstacle. 
Fig.~\ref{fig:real_world_tasks}a shows the object trajectory for a more difficult two obstacle case. However, similar to the simulation case and as expected, our method struggles to find a feasible solution for more complex configurations. The reasons are the same as the simulation case with the added issues of perception noise.

\section{Ablations and Generalization Studies}

\label{sec:ablations}
\begin{figure}[t]
    \centering
    \includegraphics[height=3.2cm]{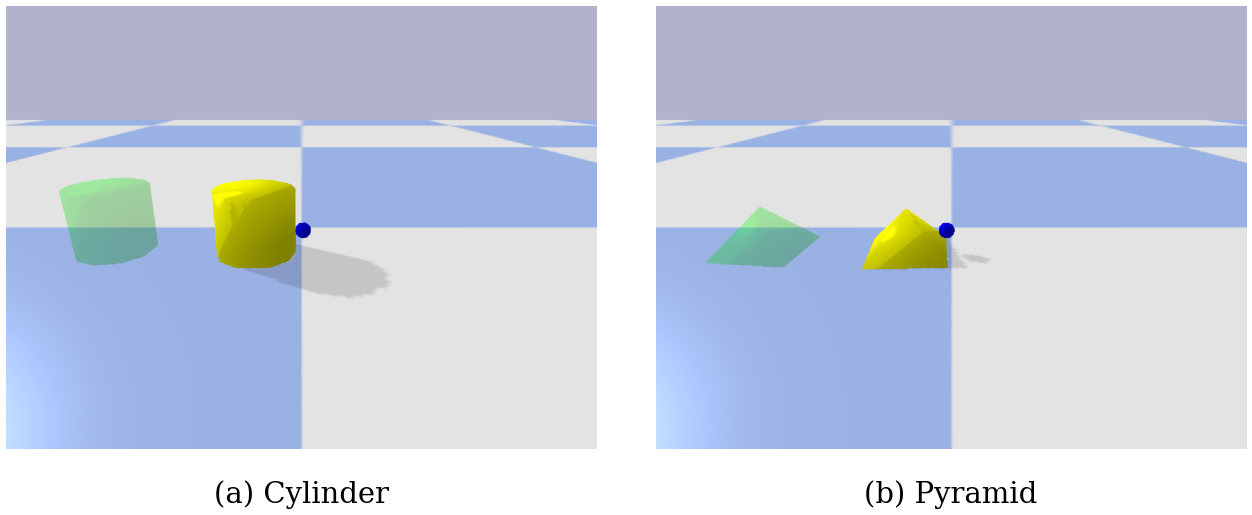}
    \caption{Unseen object shapes used to evaluate the object-specific controller. Both depart from the cube geometry used during training: (a) a cylinder and (b) a square-base pyramid with slanted faces. In each panel the solid yellow shape is the object to be pushed and the transparent green shape marks the target pose for the episode.}
    \label{fig:unseen_objects}
\end{figure}

To assess both the generality of our action selection strategy across different object shapes and the contribution of our object-specific controller to overall performance, we conduct two complementary sets of ablation and one generalization studies. The first subsection examines the role of side-switching in our controller and its impact on success rate when removed or restricted. The second subsection evaluates the object-specific controller on previously unseen object shapes, namely a cylinder and a pyramid (Fig.~\ref{fig:unseen_objects}), to assess how well the underlying push model and sampling procedure generalize beyond the cube objects used during initial training. The third subsection investigates the role of key hyperparameters and the specific modifications introduced in the object-specific controller relative to the base controller, quantifying their individual contributions to overall performance.

\subsection{Side-Switching Ablation}

A key feature that distinguishes our controller from prior sampling-based approaches, such as the one proposed by \citet{cong2020self}, is the ability to dynamically select and switch the side of the object being pushed, as well as the contact point, throughout an episode. To isolate and quantify the contribution of this capability, we design a set of ablations in which the choice of pushing side and contact point is fixed at the start of the episode and never updated thereafter, removing the controller's ability to adapt its contact strategy as the episode progresses.

We evaluate this ablation in three configurations, all of which remove the side-switching capability but differ in how the initial side and contact point are chosen and in whether the orientation cost is included in the controller's objective. In the first configuration, denoted \textit{switch}, the initial pushing side and contact point are selected by hand to be optimal for the given goal, and the episode then proceeds with this point fixed for its entire duration. In the second configuration, denoted \textit{switch 0}, the initial side and contact point are instead selected by the controller itself, and the episode similarly proceeds with this single choice fixed throughout, but with the orientation cost removed from the controller's objective. The third configuration, denoted \textit{switch 0.03}, follows the same controller-selected, single-side, single-point procedure as \textit{switch 0}, but retains the orientation cost in the controller's objective as 0.03.

Table~\ref{tab:switch_ablation} reports the success rates for each configuration across a range of position thresholds (1\,cm, 2\,cm, and 5\,cm) and orientation thresholds (5\textdegree, 10\textdegree, and 20\textdegree). For reference, our full method, which retains the ability to dynamically switch sides and contact points throughout the episode, achieves a success rate of 100\% across all of these thresholds. By contrast, even the \textit{switch} configuration, in which the initial side and point are handcrafted to be optimal, achieves at most 0.88 success at the 5\,cm threshold and only 0.32 at the 1\,cm threshold, illustrating that a single fixed contact strategy, however well chosen at the outset, is insufficient to reliably reach tight position thresholds. The \textit{switch 0} configuration, in which the controller itself selects the initial side and point but without orientation cost in its objective, performs considerably worse, with success rates as low as 0.17 at the 1\,cm threshold and never exceeding 0.61 even at the most lenient 5\,cm threshold, and similarly weak performance on the orientation thresholds (0.32, 0.33, and 0.45 at 5\textdegree, 10\textdegree, and 20\textdegree, respectively). The \textit{switch 0.03} configuration, which restores the orientation cost while still fixing the side and point after the controller's initial choice, performs worse than \textit{switch} on position thresholds, achieving a success rate of only 0.02 at 1\,cm and 0.36 at 5\,cm, but performs substantially better on the orientation-only thresholds, reaching success rates of 0.96, 1.00, and 1.00 at 5\textdegree, 10\textdegree, and 20\textdegree, respectively.

\begin{table}[tb]
\centering
\caption{Success rates for the side-switching ablation across position and orientation thresholds.}
\label{tab:switch_ablation}
\begin{tabular}{lcccccc}
\toprule
 & 1\,cm & 2\,cm & 5\,cm & 5\textdegree & 10\textdegree & 20\textdegree \\
\midrule
switch      & 0.32 & 0.66 & 0.88 & 0.85 & 0.90 & 0.93 \\
switch 0    & 0.17 & 0.33 & 0.61 & 0.32 & 0.33 & 0.45 \\
switch 0.03 & 0.02 & 0.05 & 0.36 & 0.96 & 1.00 & 1.00 \\
\bottomrule
\end{tabular}
\end{table}

These results demonstrate that the ability to dynamically switch the pushing side and contact point during an episode is essential to achieving high positional and orientational accuracy, dropping from a 100\% success rate with our full method to at most 0.88 positional and 0.93 rotational performance with generous thresholds even when the initial side and point are handcrafted to be optimal. The gap between \textit{switch} and \textit{switch 0}/\textit{switch 0.03} further shows that a handcrafted initial choice, informed by knowledge of the goal, is  preferable to letting the controller choose a single side and point on its own and then committing to it for the rest of the episode. 

\subsection{Object-Specific Controller on Unseen Objects}
The second set of ablations evaluates the generalization of our object-specific controller to different object shapes, namely cylinder and pyramid. 

We first evaluate on a cylindrical object with a radius of 6\,cm and a height of 7\,cm, standing upright on the table. Although the cylinder's size falls within the range covered by domain randomization, its shape differs fundamentally from the cube objects used to train both the underlying push model and the object-specific sampling procedure. Despite this mismatch, our method achieves a success rate of 100\% in reaching the 1\,cm position threshold. To quantify the efficiency of this generalization, we compare the number of steps required by the cylinder controller against the standard cube controller ($N=100$, push range 3\,mm--5\,cm) using Welch's $t$-tests ($n=10$ per category), separately for the short-horizon and long-horizon settings.

\textbf{Short-horizon cylinder.} At the 5\,cm threshold, we found no statistically significant difference between the cube results and the cylinder results. ($p > 0.14$ in all cases, differences of at most 0.68 steps with no consistent direction). 
At the 2\,cm threshold, the cylinder requires significantly more steps in Categories 2, 4, and 5 (differences of 1.75--2.75 steps, $p < 0.04$), while Categories 1, 3, and 6 remain non-significant ($p > 0.10$). At the 1\,cm threshold, the gap widens: Categories 1, 2, 4, 5, and 6 are all significant ($p < 0.05$, differences of 2.15--7.30 steps), with Category 2 showing the largest gap (Cyl $= 10.50 \pm 5.68$ vs.\ Std $= 3.20 \pm 0.65$, $p = 0.003$). Category 3 has some statistical trend at 1\,cm ($p \approx 0.06$).

\textbf{Long-horizon cylinder.} At the 5\,cm threshold, we did not find any statistical significance between the cube results and the cylinder results. ($p > 0.08$ in all cases, with differences of 0.35--1.05 steps), mirroring the short-horizon pattern. At the 2\,cm threshold, Categories 1, 2, 3, and 6 are significant (differences of 1.65--3.15 steps, $p < 0.04$), while Categories 4 and 5 are not ($p > 0.15$). At the 1\,cm threshold, Categories 1, 3, and 6 show highly significant differences (differences of 3.25--4.60 steps, $p < 0.005$); Categories 2 and 4 showed some statistical significance 
and Category 5 did not have significant differences in both variants ($p > 0.05$).

The results are reflecting the compounding effect of the geometry mismatch over longer episodes. In all cases the cylinder achieves 100\% success at the 1\,cm threshold, confirming that the increased step count represents a slower but reliable convergence rather than a failure of generalization.

We further evaluate on a square-base pyramid with a base side length of 6\,cm and a height of 7\,cm---a shape that departs even more radically from the training distribution, featuring four triangular slanted faces meeting at an apex rather than flat vertical sides. Crucially, we do not retrain the model or redesign the sampler; we use the same cube-trained model and cube-based sampling strategy throughout, applying only a single geometric correction: the sampled pusher contact points are shifted 3\,cm inward from their default positions to account for the reduced effective contact area of the tapered geometry. With this minimal adaptation, our controller achieves a 100\% success rate at the 1\,cm position threshold, demonstrating that the method can accommodate fundamentally non-prismatic objects with slanted surfaces provided a lightweight shape-aware correction to the contact sampling is applied. We compare against the standard cube controller in both the short-horizon and long-horizon settings using Welch's $t$-tests ($n=10$ per category).

\textbf{Short-horizon pyramid.} At the 5\,cm threshold, the pyramid and the cube controller did not show statistically significance across all six categories ($p > 0.37$ in all cases, differences of at most 0.40 steps with no consistent direction). At the 2\,cm threshold, only Category~4 reaches significance (pyramid higher by 2.35--2.45 steps, $p = 0.008$--$0.010$); the remaining five categories are non-significant ($p > 0.07$). At the 1\,cm threshold, Categories~1 and 4 are significant (pyramid higher by 2.25--2.30 and 3.35 steps respectively, $p < 0.01$), while Categories~2, 3, 5, and 6 remain non-significant ($p > 0.09$); notably, Category~6 shows the cube controller marginally higher by 0.60--0.65 steps, though this is not significant.

\textbf{Long-horizon pyramid.} At the 5\,cm threshold, the pyramid and the long-distance cube controller are statistically indistinguishable across all six categories ($p > 0.22$, differences of at most 0.80 steps). At the 2\,cm threshold, all six categories remain non-significant ($p > 0.06$, differences of 0.00--1.45 steps with no consistent direction)---a striking result given that several cylinder categories showed significant gaps at this same threshold. At the 1\,cm threshold, under wide initial sampling only Category~1 is significant (pyramid higher by 2.45 steps, $p = 0.015$); under narrow initial sampling Category~1 again reaches significance ($p = 0.015$) and Category~3 shows the cube controller higher by 1.15 steps ($p = 0.037$), while all other categories remain non-significant ($p > 0.09$).

\begin{figure}[t!]
    \centering
    \includegraphics[width=0.92\columnwidth]{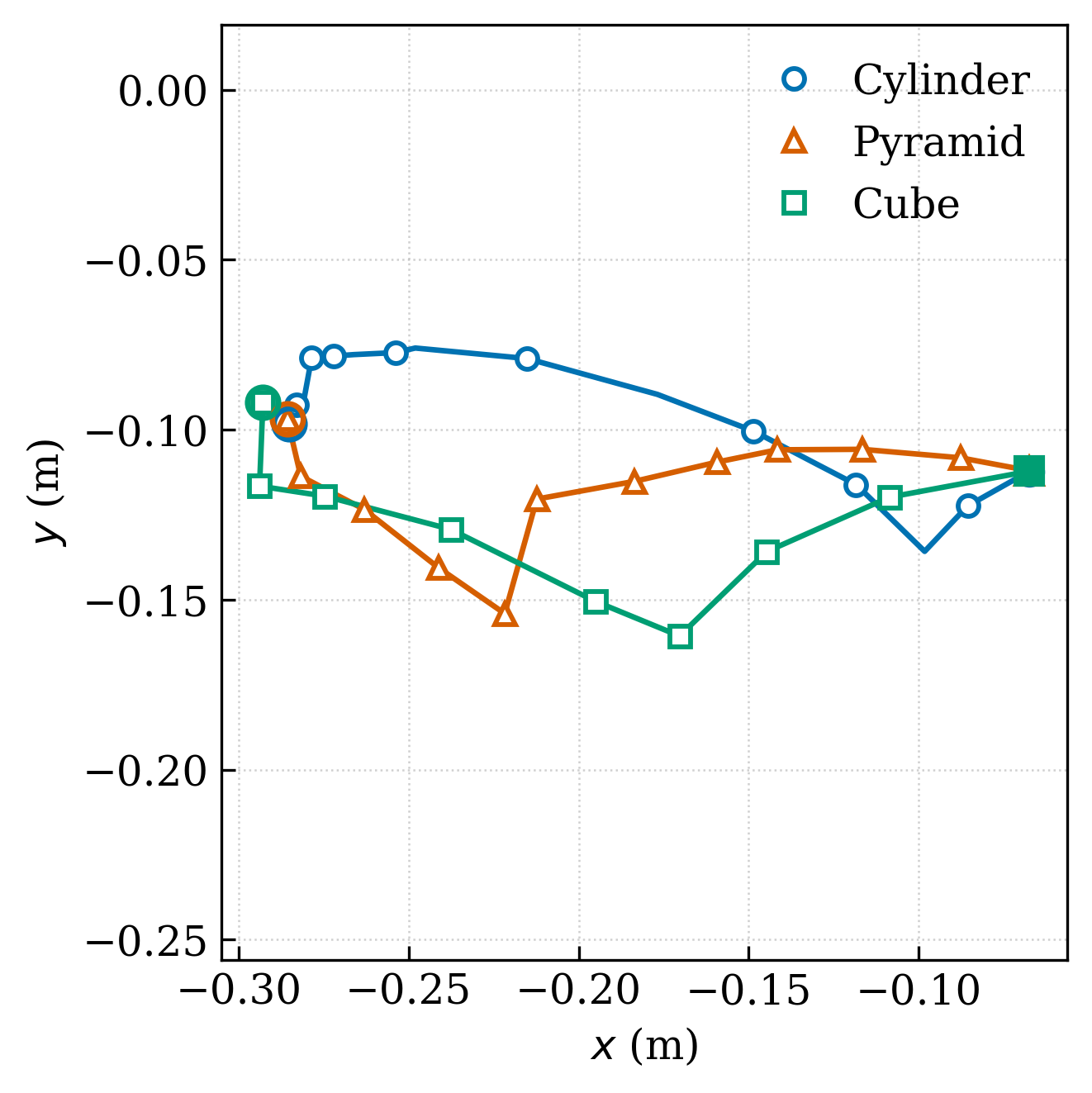}
    \caption{Representative object position trajectories under the object-specific controller for the cube and the two unseen shapes (cylinder and pyramid), driven to the same goal from the same starting point. Each marker is one push; the filled square marker is the start pose and the open ring the final pose. All three shapes reach the goal. 
    The cube and pyramid trajectories are non-linear because the controller also drives the object toward the goal orientation, trading off position and orientation corrections; the cylinder, being rotationally symmetric, has no meaningful orientation to correct, so its trajectory is comparatively straight. In addition, random sampling affects the trajectories.}
    \label{fig:unseen_trajectories}
\end{figure}

While the number of steps required to reach the goal increases relative to the cube-only setting, our method still achieves a success rate of 100\% in reaching the 1\,cm threshold for the pyramid-shaped object, similar to the cylinder case. Figure~\ref{fig:unseen_trajectories} shows representative position trajectories for the three shapes pushed to the same goal, each terminating once the object reaches the 1\,cm position threshold. All three converge, but with different paths which is due to orientation matching for cube and cylinder, and random sampling. 

As a further test of generalization, we train two additional models: one trained exclusively on pyramid-shaped objects, and one trained on a combined dataset of pyramid-shaped and cube-shaped objects under domain randomization. For both, we employ a pyramid-specific sampling strategy for data collection, adapted to sample pushing points appropriate to the pyramid's geometry. The prediction performance of both models is reported in Table~\ref{tab:pyramid_cube}. The pyramid-only model achieves an average position error of 1.3\,mm and an average orientation error of 0.012\,rad, demonstrating that a dedicated model captures pyramid push dynamics accurately. The combined model achieves 2.0\,mm and 0.015\,rad respectively, showing a modest increase in error consistent with the greater diversity of the training distribution. Our results show that our pushing model has the potential to accommodate multiple object types within a single model.
 
\begin{table}[tb]
\centering
\caption{Prediction performance of models trained on pyramid-only and combined pyramid--cube datasets, reported in millimeters for position error and radians for orientation error, with standard deviations in parentheses.}
\label{tab:pyramid_cube}
\begin{tabular}{lcc}
\toprule
Model & Avg.Pos.Err.\ (mm) & Avg.Ort.Err.\ (rad) \\
\midrule
Pyramid  & 1.3 (3.0) & 0.012 (0.050) \\
Combined      & 2.0 (2.0) & 0.015 (0.020) \\
\bottomrule
\end{tabular}
\end{table}

 \subsection{Effect of Sample Count and Push Range}
 
We further examine the sensitivity of the object-specific controller to two of its key sampling parameters: the number of sampled actions per control iteration, $N$, and the range of push magnitudes from which actions are drawn. Our standard object-specific configuration samples $N=100$ actions per iteration, distributed evenly across the four sides of the object, with push magnitudes ranging from 3\,mm to 5\,cm. We compare this standard configuration against two reduced variants, each modifying one of these parameters at a time: a \emph{narrow-range} variant that retains $N=100$ samples but restricts push magnitudes to 3\,mm--3\,cm, and a \emph{reduced-sample} variant that retains the 3\,mm--5\,cm push range but reduces the sample count to $N=20$. For each category, threshold, and horizon, we compare the standard configuration against both reduced variants and report the results for when a variant differs more from the standard configuration, treating the two as a single combined ablation of reduced sampling resources. Both reduced variants are evaluated under the same long-horizon (10--20\,cm and 20--30\,cm target distances) and short-horizon (5--15\,cm and 15--20\,cm target distances) settings used throughout our object-specific evaluation, across all six categories and four success thresholds (5\,cm, 2\,cm, 1\,cm, and 3\,mm+2\textdegree). For each comparison we report Welch's two-sample $t$-test, with $n=10$ episodes per category and configuration. 

\textbf{Long-horizon results.} At the 5\,cm threshold, the reduced configurations require substantially more steps than the standard configuration across every category, with differences ranging from approximately 1.75 to 7.1 steps and all six categories reaching statistical significance in at least one of the two reduced variants ($p<0.05$, with several categories reaching $p<0.001$). This represents the largest and most consistent gap observed across any precision level, indicating that both a wider push range and a larger sample count are most valuable when the object is still far from the goal and large, fast-progressing pushes are advantageous.
 
At the 2\,cm threshold, the gap narrows considerably and becomes inconsistent across categories. Categories 1 through 6 remain statistically significant against at least one reduced variant in most cases (differences of approximately 0.6--5.25 steps, $p$ ranging from $<0.001$ to $0.048$), with the narrow-range variant ($N=100$, 3\,mm--3\,cm) producing the larger and more consistently significant gaps, while the reduced-sample variant ($N=20$, 3\,mm--5\,cm) produces smaller, often non-significant differences in several categories (Categories 1, 2, 4, and 5, with $p$ ranging from approximately $0.14$ to $0.29$).
 
At the 1\,cm threshold, the narrow-range variant continues to show statistically significant differences from the standard configuration in nearly all categories (differences of approximately 2.55--4.50 steps, $p<0.05$ in all six categories), whereas the reduced-sample variant shows no significant differences in any category (differences of roughly 0.3--2.2 steps, $p$ ranging from approximately $0.06$ to $0.71$).
 
At the 3\,mm+2\textdegree\ threshold, the reduced-sample variant is statistically indistinguishable from the standard configuration across all six categories ($p$ ranging from approximately $0.16$ to $0.96$, differences as small as 0.05--1.85 steps with no consistent direction). The narrow-range variant shows a significant difference only in Category 5 (differences of approximately 3.20--3.85 steps, $p<0.001$), with all other categories non-significant ($p$ ranging from approximately $0.10$ to $0.85$).

\textbf{Short-horizon results.} At the 5\,cm threshold, both reduced variants are largely statistically indistinguishable from the standard configuration across most categories ($p$ ranging from approximately $0.01$ to $1.00$), with absolute differences of at most 0.65 steps for the reduced-sample variant and up to 2.05 steps for the narrow-range variant; for the reduced-sample variant, Category 3 is in fact identical to the standard configuration (1.40 vs.\ 1.40 steps, $p=1.00$). The narrow-range variant ($N=100$, 3\,mm--3\,cm) does reach significance in Categories 4--6 (differences of approximately 1.45--2.05 steps, $p<0.05$), while the reduced-sample variant ($N=20$, 3\,mm--5\,cm) shows no significant differences at this threshold. This contrasts with the long-horizon case at the same threshold, where both reduced variants produced large, highly significant increases across every category.
 
At the 2\,cm threshold, the pattern strengthens: the narrow-range variant requires significantly more steps than the standard configuration in all six categories (differences of approximately 0.90--2.65 steps, $p<0.05$ in five of six categories and $p\approx0.08$ in the remaining one), while the reduced-sample variant shows an even larger and more consistent effect, with five of six categories reaching significance (differences of approximately 1.65--3.55 steps, $p<0.01$ in most cases). Only Category 3 fails to reach significance for the reduced-sample variant (differences of 0.60--0.90 steps, $p\approx0.06$--$0.22$).
 
At the 1\,cm threshold, a similar pattern holds for the reduced-sample variant: four of six categories show statistically significant increases (differences of approximately 1.85--2.30 steps, $p$ ranging from $<0.001$ to $0.018$), while Categories 3 and 6 do not reach significance. Notably, for Category 3 the direction reverses for the reduced-sample variant: it requires slightly \emph{fewer} steps than the standard configuration (differences of $-0.75$ steps, $p\approx0.09$--$0.10$), though this difference is not statistically significant. The narrow-range variant also shows significant differences in most categories at this threshold (differences of approximately 1.25--3.85 steps, $p<0.05$ in five of six categories), with Category 6 the only exception.
 
At the 3\,mm+2\textdegree\ threshold, neither reduced variant differs significantly from the standard configuration in most categories ($p$ ranging from approximately $0.10$ to $0.77$ for the reduced-sample variant, and similarly non-significant for most categories of the narrow-range variant aside from isolated exceptions), with differences of up to 2.45 steps but no consistent direction across categories.
 
Taken together, these results show that the impact of restricting the push-magnitude range to 3\,mm--3\,cm (at $N=100$) or reducing the sample count to $N=20$ (at a 3\,mm--5\,cm range) is highly dependent on both the horizon length and the strictness of the success threshold, and the two horizons do not exhibit the same trend. These results imply that our original parameters are better in general but that there is potential in adjusting the sample counts and/or the push length dynamically based on the required precision and target distance.

\section{Discussion}
In this work, we proposed a framework that decouples the dynamics model from the controller for planar pushing. 
Our dynamics model achieved high accuracy: less than $1\,mm$ 
and $0.5^\circ$ error on a basic dataset, and less than $2\,mm$ 
and $2^\circ$ error on a more diverse dataset. 
Considering the variety in pushing lengths and directions, as well as the consistent performance observed in the precision positioning task, these results demonstrate that our model successfully learns the dynamics of cubic object pushes and generalizes to new environments and object sizes with our representation.

A detailed comparison with other methods is not straightforward, since many of them incorporate additional objects and perception-based state representations, making their models inherently aware of object shape and size. On the flip side, these methods do not employ stringent success thresholds and do not evaluate their approach on multiple tasks. To the best of our knowledge, one of the most directly comparable works is \citet{10802843}, which also demonstrates side-switching behavior; however, their system is designed exclusively for trajectory following and primarily evaluates model predictions rather than full pushing-to-goal performance. Furthermore, their framework does not address obstacle avoidance, and their real-world trajectory-following results are comparable to ours, despite our system tackling a broader set of tasks. While prior works such as \citet{bauza2017probabilistic, 10802843} employ non-parametric methods for learning forward dynamics models, our approach leverages a recurrent neural network to incorporate history encoding and to enable faster prediction. The proposed model demonstrated better object-specific performance compared to the method presented in \citet{cong2020self}. Specifically, we incorporated additional MLP layers at the final stages of the GRU-based architecture, enabling the model to better capture the non-linear dynamics inherent in pushing interactions. Furthermore, by including the relative pose in our state representation in addition to the push magnitude, we enabled side-switching behavior during execution. Side-switching allows the controller to adaptively change the pushing side and select optimal contact points to guide the object more efficiently toward the target pose. A critical design aspect that enabled this capability was the computation of relative pose and push magnitudes with respect to the object’s current orientation, ensuring consistency and generalization across varying scenarios.

Our extensive experiments demonstrate that the proposed method can achieve sub-centimeter accuracy and reach within 2~cm of the goal pose in under 10 steps, even for long-horizon tasks spanning distances of 20--30~cm. This represents a significant improvement over prior non-prehensile manipulation methods, which typically aimed only to push the object from its initial pose to the target pose without explicitly handling precise intermediate positioning. We improved the framework by extending the push length, incorporating balanced sampling across sides, and introducing greedy action selection on top of the base MPPI controller. These enhancements effectively reduced the number of steps required to reach the goal and lowered the overall motion planning effort of the robot. Visual and quantitative comparisons between the basic and object-specific versions clearly show the benefits: the object-specific framework achieves faster convergence while balancing position and orientation corrections, particularly in longer-distance tasks where greater positional freedom allows for better orientation adjustments.

Overall, our results highlight significant improvements over previous model-based approaches while, at the same time, demonstrating advantages over the task-specific model-free approaches considered here. Unlike model-free strategies, which often require task-specific training, our learned model can be adapted to multiple tasks simply by modifying the cost function or control objective, without additional retraining. This flexibility underscores the broader potential of combining model-based learning with sampling-based control for robust and versatile non-prehensile manipulation. We argue that our state-action representation, model inductive bias and other sampling based modifications synergistically result in a flexible and robust pushing system that can trade off task speed and precision.

\section{Limitations}


Our results are subject to several qualifications. First, the object-size range is deliberately narrow. Larger geometric variation changes the effective contact geometry, which can increase model uncertainty and degrade prediction accuracy. Our results therefore establish generalization within a controlled range of sizes rather than across arbitrary scales. Second, generalization to non-prismatic shapes is not fully geometry-agnostic. The cube-trained dynamics model transfers to the square-base pyramid without retraining, but reliable performance required a lightweight shape-aware correction: because the pyramid’s tapered faces alter the effective contact geometry, the sampled contact points were shifted inward to ensure contact. 

Third, our method is formulated for quasi-static planar pushing. This is a common assumption in the field and has been characterized as ubiquitous in much of the planar-pushing literature~\cite{10.3389/frobt.2020.00008}. In our real-world experiments, we cap the pusher speed at 10 cm/s to remain within this operating regime. Accordingly, pusher velocity and action duration are not included as independent inputs to the learned dynamics model, and we do not claim generalization to dynamic pushing regimes in which inertial or speed-dependent effects become significant. We have not characterized sensitivity to pushing speed or the boundary at which the quasi-static approximation ceases to be adequate. 

Finally, the controller relies on sampling-based rollouts, while the real-world system relies on externally estimated object poses obtained through markerless tracking. The learned dynamics model operates on pose-level state information rather than raw perceptual or explicit shape features, which limits scalability to richer environments and more varied geometries. Broader size and shape generalization, automatic shape-aware sampling, richer perceptual representations, and sensitivity analysis with respect to pushing speed are natural directions for future work.

\section{Conclusion and Future Work}
We propose a novel system for planar pushing that builds upon a GRU-based dynamics model combined with a sampling-based controller. While similar frameworks have been proposed before, what makes our approach unique is its adaptability to multiple objectives, its capability for high-precision manipulation, and its ability to switch pushing sides using pose estimates obtained through markerless tracking, without requiring raw visual input to the dynamics model. The dynamics model is trained in simulation using extensive domain randomization to improve generalization to real-world conditions. By leveraging a state–action representation encoding both relative pose and push directions, our system enables versatile capabilities such as object-side switching, variable-length pushes, and the integration of diverse task objectives. This integrated framework effectively exploits the learned dynamics to generate adaptive, task-oriented actions, achieving not only precise positioning but also behaviors like trajectory following and obstacle avoidance. Through systematic evaluation, we demonstrated the efficacy of our method in both simulated environments and real-world experiments with a Franka Panda robot equipped with markerless tracking. The results confirmed high success rates in precise positioning tasks under demanding position and orientation thresholds, while also showing robust performance in other tasks like obstacle avoidance and path following. Building upon this foundation, we also introduced an object-specific version of our framework designed to address long-horizon tasks more efficiently. Specifically, we trained a model capable of handling longer pushes and incorporated a balanced sampling strategy across object sides. Combined with a greedy action selection mechanism, this enhancement allowed the controller to prioritize high-quality actions, significantly reducing the number of steps required to achieve longer-horizon objectives while retaining precision and robustness.

Despite these encouraging results, several avenues remain for improvement. Our current controller, although enhanced with balanced sampling, still relies on sampling-based rollouts, which can be computationally demanding. Future work can explore more sophisticated goal-aware sampling strategies, for example by learning a sampling policy or developing an inverse dynamics model to guide exploration. Additionally, learning a dedicated sampling function could streamline the selection of optimal actions within rollouts, thereby reducing planning time while improving efficiency. These enhancements would allow the framework to scale to more complex environments, enabling it to handle more challenging obstacle-avoidance and long-horizon planning tasks. Another promising direction is to integrate our method with a global planner, allowing the system to seamlessly combine local pushing strategies with long-range planning. Finally, to broaden the applicability of our approach, future efforts will focus on integrating perceptual representations, which would allow the model to generalize across a wider variety of object geometries.

\section*{Declarations}
\noindent \textbf{Funding} \\
This project was supported by the KUIS AI Lab at Koç University. Aydin Ahmadi contributed to this work as a research fellow at the KUIS AI Lab.

\noindent \textbf{Conflict of interest} \\
The authors declare no conflicts of interest.

\noindent \textbf{Author contribution} \\
Aydin Ahmadi developed the forward dynamics model, the model predictive control framework, and the simulation environment. He performed both the simulation experiments and the real-world experiments on the Franka Emika Panda robot. Barıs Akgun served as his advisor and guided the project throughout his Master’s studies.


\bibliography{sn-bibliography.bib}

\end{document}